\documentclass[authoryear,final,3p,11pt]{elsarticle}

\usepackage{amsmath,bm}
\usepackage{bbm}
\usepackage{url}
\usepackage{tabularx}
\usepackage{colortbl,booktabs}

\usepackage{booktabs}
\usepackage{multirow}
\usepackage{threeparttable}
\usepackage{algorithm}
\usepackage{verbatim}
\usepackage{algpseudocode}
\usepackage{xcolor}
\usepackage{diagbox}
\usepackage{enumerate}
\usepackage{threeparttable}
\usepackage{graphicx}
\usepackage{subfigure}
\usepackage{algpseudocode}
\usepackage{makecell}
\usepackage{nicematrix}
\usepackage{hyperref}
\usepackage{eurosym}
\usepackage{siunitx}
\hypersetup{
    colorlinks=true,
    linkcolor=blue,
    filecolor=blue,      
    urlcolor=blue,
}

\usepackage{amsthm,amsmath,amssymb}
\usepackage{tablefootnote}

\usepackage{hyperref}

\usepackage[toc,page]{appendix}
\usepackage{setspace}
\theoremstyle{definition}
\newtheorem{proposition}{Proposition}[section]
\newtheorem{remark}{Remark}[section]

\theoremstyle{definition}
\newtheorem{definition}{Definition}[section]

\theoremstyle{definition}
\newtheorem{assumption}{Assumption}[section]

\theoremstyle{definition}
\newtheorem{theorem}{Theorem}[section]

\theoremstyle{definition}
\newtheorem{corollary}{Corollary}[section]

\journal{European Journal of Operational Research}

\begin{document}

\begin{frontmatter}



\title{Value-oriented forecast reconciliation for renewables in electricity markets\vspace{-0.75em}}


\author[inst1,inst2]{Honglin Wen\corref{cor1}}
\ead{linlin00@sjtu.edu.cn}
\cortext[cor1]{Corresponding authors}

\author[inst2,inst3,inst4,inst5]{Pierre Pinson\corref{cor1}}
\ead{p.pinson@imperial.ac.uk}

\affiliation[inst1]{organization={Department of Electrical Engineering, Shanghai Jiao Tong University, China}}

\affiliation[inst2]{organization={Dyson School of Design Engineering, Imperial College London, United Kingdom}}

\affiliation[inst3]{organization={Halfspace, Denmark}}

\affiliation[inst4]{organization={Department of Technology, Management and Economics, Technical University of Denmark, Denmark}}

\affiliation[inst5]{organization={CoRE, Aarhus University, Denmark}}


\begin{abstract}
Forecast reconciliation is considered an effective method to achieve coherence (within a forecast hierarchy) and to improve forecast quality. However, the value of reconciled forecasts in downstream decision-making tasks has been mostly overlooked. In a multi-agent setup with heterogeneous loss functions, this oversight may lead to unfair outcomes, hence resulting in conflicts during the reconciliation process. To address this, we propose a value-oriented forecast reconciliation approach that focuses on the forecast value for all individual agents. Fairness is ensured through the use of a Nash bargaining framework. Specifically, we model this problem as a cooperative bargaining game, where each agent aims to optimize their own gain while contributing to the overall reconciliation process. We then present a primal-dual algorithm for parameter estimation based on empirical risk minimization. From an application perspective, we consider an aggregated wind energy trading problem, where profits are distributed using a weighted allocation rule. We demonstrate the effectiveness of our approach through several numerical experiments, showing that it consistently results in increased profits for all agents involved.

\end{abstract}




\end{frontmatter}


\section{Introduction}
\label{s1}
The geographically distributed nature of renewable energy sources, such as wind and solar, inherently creates a hierarchical structure for power generation. 
It is acknowledged that forecast accuracy can be improved by leveraging the spatial-temporal dependencies inherent in geographically distributed data \citep{tastu2013probabilistic}. However, various agents at different levels of aggregation in power systems and electricity markets, for instance, the portfolio and individual power generation assets, often independently generate their own forecasts. This can lead to \textit{incoherence}, i.e., the forecast directly produced at the portfolio level is not necessarily equal to the sum of the forecasts for individual assets. Given that forecasts are frequently used in decision-making, incoherence across aggregation levels may eventually lead to dilemmas among decision-makers. For that, \textit{forecast reconciliation} is applied, which is a post-forecasting process that involves transforming independently generated incoherent base forecasts into coherent forecasts that satisfy the constraints imposed by the hierarchical structures \citep{athanasopoulos2023forecast}.

Single-level reconciliation approaches, such as bottom-up, top-down, and middle-out approaches, have been in use for decades, dating back at least to \citet{edwards1969should} and \citet{zellner1969aggregation}. They require forecasts at a chosen level, which are then combined to generate forecasts for the remaining levels in the hierarchy. \citet{hyndman2011optimal} introduced a combination approach that leverages forecasts across the entire hierarchy. This was later extended by \citet{van2015game}, \citet{hyndman2016fast}, and \citet{wickramasuriya2019optimal} through employing various projection matrices \citep{panagiotelis2021forecast}. In parallel, \citet{ben2019regularized} developed an end-to-end reconciliation approach, which learns the projection matrices via machine learning. Forecast reconciliation in the energy sector has been employed since \citet{van2015game}, and most often brings the added benefit of enhanced forecast accuracy \citep{zhang2018least,jeon2019probabilistic}.

While previous studies have substantially advanced forecast reconciliation through various methods and techniques, nearly all of them focus on leveraging hierarchical structures to improve forecast accuracy. Nonetheless, as \citet{murphy1993good} suggests, in addition to evaluating the quality (accuracy) of forecasts, which is the correspondence between forecasts and observed outcomes, it is also important to assess the consistency, or the alignment between forecasts and judgments, and the value, which refers to the benefits they provide to users, in terms of profits or costs. Evidence suggests that more accurate forecasts do not always lead to higher value in downstream decision-making processes \citep{kourentzes2013intermittent,barrow2016distributions}. Although increasing emphasis is placed on developing forecast products that maximize value in decision-making, particularly within the energy industry \citep{carriere2019integrated, stratigakos2022prescriptive, zhang2023value} and inventory management \citep{kourentzes2020optimising,theodorou2025forecast}, especially after the theoretical analysis by \citet{bertsimas2020predictive} and \citet{elmachtoub2022smart}, value in forecast reconciliation remains largely under-investigated, with only a few notable exceptions. For instance, \citet{sanguri2022hierarchical} assessed the effectiveness of coherent forecasts in enhancing the value of port operations and empirically demonstrated that coherent forecasts improved decision-making at ports. In parallel, \citet{stratigakos2024decision} proposed a method for combining multiple probabilistic forecasts using a linear opinion pool, with a particular emphasis on minimizing the expected costs associated with subsequent decision-making. However, all of these studies evaluate the value in forecast reconciliation from a centralized perspective, neglecting the fact that the agents involved may have heterogeneous loss (or cost) functions. 

When agents have heterogeneous cost functions, the primary challenge lies in achieving consensus on the reconciliation approach before actually reconciling their forecasts. Since the commonly used reconciliation approaches aim to optimize an overall loss function, for instance, the mean squared error function \citep{van2015game, hyndman2016fast}, some agents involved may even find that reconciled forecasts lead to an increase in their own costs, even though overall costs are reduced after reconciliation. As a result, the agents may perceive the reconciliation process as unfair, leading to disagreements on the reconciliation procedure. Indeed, studies like \citet{van2015game} and \citet{sprangers2024hierarchical} included weighting factors for forecasts at each level during the reconciliation process, either on an ad hoc basis or based on hierarchical levels. Nevertheless, the emphasis is primarily placed on the overall accuracy of the forecasts, thereby leaving the effects of this weighting on subsequent decision-making results unexplored. Therefore, how to account for the value of forecasts at each level and ensure fairness in the reconciliation process remains an unresolved issue in the forecast reconciliation literature. This paper addresses this gap by introducing a \textit{value-oriented reconciliation} approach, which we contrast with the widely studied \textit{quality-oriented reconciliation} approach.


To motivate our proposal and demonstrate its applicability, we focus on the well-established problem of trading wind energy in electricity markets, where a portfolio manager (PM) trades on behalf of multiple wind power producers (WPPs). In this setup, a natural two-level hierarchy emerges between the PM and the WPPs, which implies that the PM's offer must represent the collective output of all WPPs. As the forecasts issued by the PM and WPPs may be incoherent, they must be reconciled to ensure coherence before being used in the PM's trading strategies. After trading in the markets, the PM allocates profits to each WPP by using pre-defined allocation rules.

Our reconciliation framework is inspired by \citet{pennings2017integrated}, where the reconciled forecasts are expressed as the product of a structural matrix (defined by the hierarchy) and latent variables. The latent variables are derived from the base forecasts via learnable combination functions. Market data, including historical prices, are incorporated as context features for the combination functions, enabling the reconciliation to be adaptive to contextual information. 
The requirement for consensus on the reconciliation approach aligns with the group decision-making process, in which agents collaboratively select a combination function from the collection of mutually accepted functions. Particularly, we assume that WPPs exhibit \textit{individual rationality}, indicating that each producer's profit must be at least equal to their independent offering. Therefore, the forecast reconciliation process must ensure that each producer’s profit is either maintained or improved after reconciliation. To address this, we model the problem as a cooperative bargaining game \citep{nash1953two}, where each agent seeks to maximize their own gain by choosing from the available combination functions. The Nash bargaining solution is derived by maximizing the product of excess profits (known as the Nash product), which measures the difference between the post-reconciliation profits and the pre-reconciliation profits.

Given that the PM regularly engages in market trading, we apply the principle of fairness over time \citep{lodi2024framework} to determine the combination function in the reconciliation process, rather than relying on a single trade instance. Consequently, we calculate the average profit for each agent within the Nash bargaining framework. The combination function is derived through constrained optimization, with the objective of maximizing the product of average excess profits while ensuring that each agent’s average excess profit remains positive. It turns out to be a large-scale optimization problem, thus we frame it as a constrained learning problem by parameterizing the combination function \citep{donini2018empirical}. Specifically, we use the Lagrangian optimization approach outlined by \citet{chamon2020probably} for parameter estimation. The parameters of the combination function model are estimated using the stochastic gradient descent algorithm, while the dual variables are updated through the dual ascent algorithm. Finally, following \citet{munoz2023online}, we validate the proposed method with case studies in an hourly forward market coupled with a dual-price settlement balancing market, using market data from the Danish TSO and wind power data from the Wind Toolkit. In summary, this work makes the following contributions:
\begin{enumerate}[(1)]
    \item We propose a value-oriented forecast reconciliation approach, with a particular emphasis on the value of forecasts for each agent involved. The approach ensures fairness through a Nash bargaining framework. It complements the existing forecast reconciliation literature by emphasizing forecast value rather than solely focusing on forecast quality.
    \item We demonstrate the effectiveness of the proposed method in the context of aggregated wind energy trading and illustrate how consensus can be achieved when dealing with forecasts from multiple sources. This extends the existing literature on aggregated wind energy trading by introducing the concept of consensus on information, in addition to the traditional focus on consensus in profit allocations.
    \item We prove in Corollary~\ref{corollary2} that, by employing weighted proportional allocation rules, the agreement set of parameters in the Nash bargaining problem—within the context of aggregated wind energy trading—is guaranteed to be non-empty.
    \item We propose an algorithm for parameter estimation within the Nash bargaining framework that alternates between maximizing over the dual variables and minimizing over the parameters of the reconciliation function.
\end{enumerate}
The structure of the paper is as follows. Section \ref{s3} introduces the core concepts of value-oriented forecasting and decision-making, with a particular emphasis on wind energy trading, and discusses the challenges associated with aggregated wind energy trading. Section \ref{s4} formulates the problem and develops a value-oriented forecast reconciliation approach based on Nash bargaining, framing it as a learning problem. Section \ref{s5} presents two illustrative cases to explore the behavior of the proposed approach. Section \ref{s6} empirically validates the approach by constructing an hourly forward market using real market data. Finally, Section \ref{s7} concludes the paper and suggests directions for future research.

\textbf{Notations:} Vectors are denoted by bold lowercase letters (e.g., $\mathbf{x}$), while matrices are represented by bold uppercase letters (e.g., $\mathbf{A}$). Sets are denoted using calligraphic fonts, such as $\mathcal{T}$, with the cardinality of a set expressed as $|\mathcal{T}|$. The subscript $\mathrm{F}$ represents the forward market (e.g., $\pi^\mathrm{F}$), while the superscripts $\mathrm{UP}$ and $\mathrm{DW}$ denote up- and down-regulation (e.g., $\pi^\mathrm{UP}$ and $\pi^\mathrm{DW}$).

\section{Preliminaries}
\label{s3}
Our setup closely mirrors that of \citet{guerrero2015optimal}. Specifically, we consider two trading floors: a forward market and a balancing market with a dual-price settlement mechanism for imbalances. As noted by \citet{pinson2007trading}, the trading problem for a risk-neutral, price-taker WPP is analogous to a newsvendor problem. Ultimately, it involves offering on a specific quantile of the predictive distribution for wind power generation. We assume that both the PM and WPPs are risk-neutral and act as price-takers, enabling us to model their interaction as a cooperative newsvendor game~\citep{montrucchio2012cooperative}. In what follows, we first outline the fundamentals of value-oriented forecasting and decision-making, with a specific focus on wind energy trading. We then present renewable energy offering in the forward market within the context of dual-price imbalance settlement. Finally, we describe the challenges associated with aggregated wind energy trading.

\subsection{Value-oriented Forecasting and Decision Making}

To begin, we outline the setups of wind energy offering. Let $y_{t,i}$ represent the wind energy generated by WPP $i$ at time $t$, which is a realization of the random variable $Y_{t,i}$. In a forward market, WPP $i$ must declare the energy amount $y_{t+k,i}^\mathrm{F}$ it will generate before the market's closure time, $t$. Here, the subscript $k$ indicates the time interval between trading and actual power delivery. In the balancing market, the WPP is required to balance the discrepancy between the actual generation, $y_{t+k,i}$, and the declared amount, $y_{t+k,i}^\mathrm{F}$. Throughout this paper, we assume that WPPs act as price takers and are risk-neutral.
This assumption is commonly observed in the electricity market literature \citep{morales2013integrating}, implying that the trading activities of a single WPP do not influence the clearing price. The prices for the forward market, upward and downward regulation are denoted by $\pi^\mathrm{F}_{t+k}$, $\pi^\mathrm{UP}_{t+k}$, and $\pi^\mathrm{DW}_{t+k}$, respectively. It is established that $\pi^\mathrm{UP}_{t+k}\geq\pi^\mathrm{F}_{t+k}$ and $\pi^\mathrm{DW}_{t+k}\leq\pi^\mathrm{F}_{t+k}$, with at most one of these values differing from $\pi^\mathrm{F}_{t+k}$. 

The profit of a WPP at the trading hour $t+k$ is then given by the function $\rho_{t+k}(y_{t+k,i}^\mathrm{F},y_{t+k,i})$, which is expressed as
\begin{equation}
\label{profit_ind}
    \rho_{t+k}(y_{t+k,i}^\mathrm{F},y_{t+k,i})=\pi^\mathrm{F}_{t+k}y_{t+k,i}^\mathrm{F}-\pi^\mathrm{UP}_{t+k}\left[y_{t+k,i}^\mathrm{F}-y_{t+k,i}\right]^++\pi^\mathrm{DW}_{t+k}\left[y_{t+k,i}-y_{t+k,i}^\mathrm{F}\right]^+,
\end{equation}
where $[x]^+=\max(x,0), \forall x\in \mathbb{R}$. When the power generation falls short of the offer, i.e., $y_{t+k,i}<y_{t+k,i}^\mathrm{F}$, the WPP has to purchase up-regulation at the price of $\pi^\mathrm{UP}_{t+k}$ to balance the discrepancy $y_{t+k,i}^\mathrm{F}-y_{t+k,i}$. Conversely, if the power generation exceeds the offer, i.e., $y_{t+k,i}^\mathrm{F}>y_{t+k,i}$, the WPP is required to purchase down-regulation at the price of $\pi^\mathrm{DW}_{t+k}$. The penalties associated with overproduction and underproduction are calculated as follows: for overproduction,
$\psi^+_{t+k} = \pi^\mathrm{F}_{t+k} - \pi^\mathrm{DW}_{t+k}$,
and for underproduction,
$\psi^-_{t+k} = \pi^\mathrm{UP}_{t+k} - \pi^\mathrm{F}_{t+k}$.
For mathematical convenience, we recast \eqref{profit_ind} as
\begin{equation}
\label{profit_ind_2}
    \rho_{t+k}(y_{t+k,i}^\mathrm{F},y_{t+k,i})=\pi^\mathrm{F}_{t+k}y_{t+k,i}-\left[ \psi^+_{t+k}\left[y_{t+k,i}-y_{t+k,i}^\mathrm{F}\right]^++\psi^-_{t+k}\left[y_{t+k,i}^\mathrm{F}-y_{t+k,i}\right]^+\right],
\end{equation}
given that
\begin{align*}
    y_{t+k,i}-y_{t+k,i}^\mathrm{F}=\left[y_{t+k,i}-y_{t+k,i}^\mathrm{F}\right]^+-\left[y_{t+k,i}^\mathrm{F}-y_{t+k,i}\right]^+.
\end{align*}
The first term on the RHS of \eqref{profit_ind_2} represents the profit derived from the oracle forecast (i.e., an ideal forecast), and consequently, it is outside the producer's influence. The subsequent term on the RHS of \eqref{profit_ind_2} is identified as the \textit{imbalance cost}, which is a non-negative quantity represented by a function $c_{t+k}(y_{t+k,i}^\mathrm{F},y_{t+k,i})$ as follows:
\begin{equation}
    c_{t+k}(y_{t+k,i}^\mathrm{F},y_{t+k,i})=\psi^+_{t+k}\left[y_{t+k,i}-y_{t+k,i}^\mathrm{F}\right]^++\psi^-_{t+k}\left[y_{t+k,i}^\mathrm{F}-y_{t+k,i}\right]^+.
    \label{imbalance cost}
\end{equation}
And we note that $\psi^+_{t+k}$ and $\psi^-_{t+k}$ are non-negative and bounded.

It is important to note that the optimal offer can be determined by minimizing the expected imbalance cost, provided that the distribution of $Y_{t+k,i}$ is known. It is expressed as
\begin{equation}
    y_{t+k,i}^{\mathrm{F}*}=\underset{y_{t+k,i}^\mathrm{F}\in[0,u_i]}{\arg\min}\ \mathbb{E}_{Y_{t+k,i}}\left[ c_{t+k}(y_{t+k,i}^\mathrm{F},y_{t+k,i})\right],
\end{equation}
where $u_i$ represents the capacity of WPP $i$. This addresses a specific instance of the newsvendor problem \citep{choi2012handbook}, which leads to the identification of a particular quantile within the wind power generation distribution:
\begin{equation}
    y_{t+k,i}^{\mathrm{F}*}=F^{-1}_{y_{t+k,i}}(\alpha_{t+k}),
    \label{optoffer}
\end{equation}
where $F_{y_{t+k,i}}$ represents the cumulative distribution function (CDF) of $Y_{t+k,i}$ and $\alpha_{t+k}$ signifies the nominal level, expressed as:
\begin{align*}
    \alpha_{t+k} = \frac{\psi^+_{t+k}}{\psi^+_{t+k}+\psi^-_{t+k}}.
\end{align*}
Thus, the expense associated with the imbalance is represented by $c_{t+k}(y_{t+k,i}^{\mathrm{F}*}, y_{t+k,i})$, and the expected cost is obtained as $\mathbb{E}_{Y_{t+k,i}}[c_{t+k}(y_{t+k,i}^{\mathrm{F}*},y_{t+k,i})]$.

However, both the real generation $y_{t+k,i}$ and its distribution $F_{y_{t+k,i}}$ are unknown at the time of offering $t$, thus WPPs must rely on forecasts to determine their offers. It is common to use the contextual information available at time $t$ such as weather and lagged values, denoted as $\mathbf{x}_{t,i}$, to predict the future value $Y_{t+k,i}$ using a specific model $f_i$. Denoting the point forecast as $\hat{y}_{t+k,i|t}$, forecasting can be mathematically expressed as
\begin{equation}
    \label{forecast_offer} \hat{y}_{t+k,i|t}=f_i(\mathbf{x}_{t,i};\phi_i),
\end{equation}
where $\phi_i$ represents the model parameters. According to the literature on electricity markets, two main point forecasting products are identified for trading purposes: mean and quantiles. They are developed by employing various loss functions to train the model (\ref{forecast_offer}), as elaborated below.

\textbf{Quality-oriented forecasting}: The widely employed point forecasting seeks to estimate the expected value of $Y_{t+k,i}$. For that, the parameters of the model are estimated by minimizing the expectation of the mean squared error function $l(\hat{y}_{t+k,i|t},y_{t+k,i})$, expressed as
\begin{equation}
    \underset{\phi_i}{\min} \ \mathbb{E}\left[ l(\hat{y}_{t+k,i|t},y_{t+k,i})\right].
\end{equation}
We refer to this approach as quality-oriented forecasting since the loss function is directly related to the quality (or accuracy) of the forecasts.

\textbf{Value-oriented forecasting}: As various studies indicate that more precise forecasts do not necessarily enhance value in subsequent decision-making \citep{theodorou2025forecast}, we can create forecasting models that directly target decision-making value. Then, the parameters of the forecast model are estimated by minimizing the expected cost, expressed as
\begin{equation}
    \underset{\phi_i}{\min} \ \mathbb{E}\left[ c_{t+k}(\hat{y}_{t+k,i|t},y_{t+k,i})\right],
\end{equation}
where $c_{t+k}(\cdot)$ is defined in \eqref{imbalance cost}.
Particularly for the wind power offering problem, \citet{bremnes2004probabilistic} and \citet{pinson2007trading} suggest that, for a specified nominal quantile level, it is allowed to directly estimate the optimal offers using quantile regression. Specifically, they employ the long-term mean values of $\psi^+_{t+k}$ and $\psi^-_{t+k}$ during the model training process. It is worth mentioning that the loss function is not limited to the pinball loss; it can also be a more generalized form as explored in studies by \citet{kourentzes2020optimising, stratigakos2022prescriptive, zhang2023value}, which address issues related to optimization or simulation. Since the loss function aligns with the value of downstream decisions, we refer to this approach as value-oriented forecasting.

\subsection{Aggregated Wind Energy Trading}
\label{s3.2}

Leveraging the complementary nature of geographically diverse renewable energy sources to hedge against price and quantity uncertainty in electricity markets has attracted significant interest \citep{bitar2011selling}.
Consider a scenario where there are $m$ WPPs, managed by a PM responsible for trading in the market on behalf of all the WPPs. Let $Y_{t+k,\mathrm{sum}}$ denote the aggregate sum of the random variables $Y_{t+k,i}$, expressed as $Y_{t+k,\mathrm{sum}} = \sum_i^m Y_{t+k,i}$. The observed value is indicated by $y_{t+k,\mathrm{sum}} = \sum_i^m y_{t+k,i}$. Meanwhile, $y^\mathrm{F}_{t+k,\mathrm{sum}}$ stands for the trading offer by the PM, and $c_{t+k}(y_{t+k,\mathrm{sum}}^\mathrm{F},y_{t+k,\mathrm{sum}})$ denotes the total imbalance cost associated with the offer $y^\mathrm{F}_{t+k,\mathrm{sum}}$. When the generation assets are owned by the same company, the portfolio manager (PM) aims to minimize the total expected imbalance cost. In this case, the problem reduces to a classical newsvendor formulation, and the optimal offer for the aggregated portfolio is given by
\begin{equation}
    y_{t+k,\mathrm{sum}}^{\mathrm{F}*}= F^{-1}_{y_{t+k,\mathrm{sum}}}(\alpha_{t+k}),
\end{equation}
where $F_{y_{t+k,\mathrm{sum}}}$ denotes the CDF of the aggregated random variable $Y_{t+k,\mathrm{sum}}$. The imbalance cost associated with the offer $y_{t+k,\mathrm{sum}}^{\mathrm{F}*}$ is then given by $c_{t+k}(y_{t+k,\mathrm{sum}}^{\mathrm{F}*},y_{t+k,\mathrm{sum}})$. Accordingly, the expected total imbalance cost is computed as $\mathbb{E}_{Y_{t+k,\mathrm{sum}}}[c_{t+k}(y_{t+k,\mathrm{sum}}^{\mathrm{F}*},y_{t+k,\mathrm{sum}})]$. 

The cost-saving benefits of aggregated trading—originally demonstrated by \citet{eppen1979note} and further supported by \citet{hartman2000cores}—are outlined below.
\begin{theorem}
\label{theorem3.1}
[\citet{eppen1979note,hartman2000cores}]
In a multi-location newsvendor system, the total optimal expected cost of the centralized system is lower than that of the decentralized system.
\end{theorem}
\noindent Using Theorem \ref{theorem3.1}, it is evident that
\begin{equation}
\label{cost_saving}
    \mathbb{E}_{Y_{t+k,\mathrm{sum}}}[c_{t+k}(y_{t+k,\mathrm{sum}}^{\mathrm{F}*},y_{t+k,\mathrm{sum}})]\leq \sum_i^m \mathbb{E}_{Y_{t+k,i}}[c_{t+k}(y_{t+k,i}^{\mathrm{F}*},y_{t+k,i})],
\end{equation}
which implies that offering as an aggregation can always lead to a reduction in costs for participants. Equality in (\ref{cost_saving}) holds only when the variables $Y_{t+k,1},\cdots,Y_{t+k,m}$ are perfectly positively correlated. 

\begin{figure}[ht]
    \centering
    \includegraphics[width=0.8\linewidth]{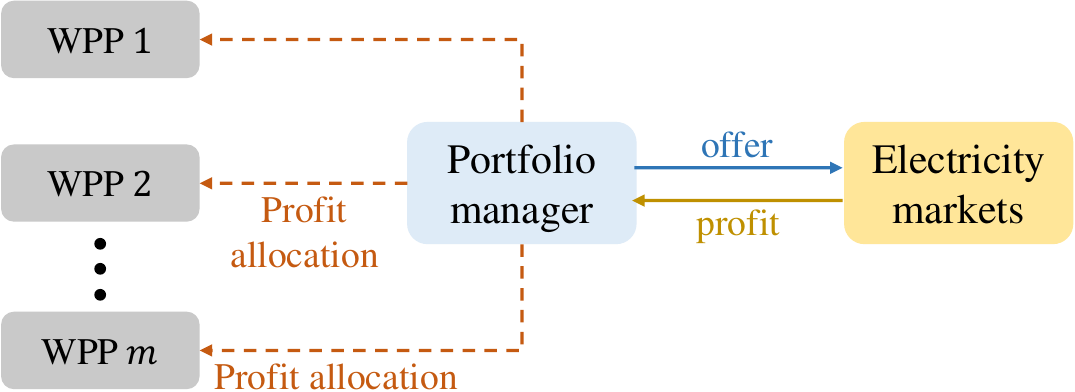}
    \caption{wind energy trading as an aggregation.}
    \label{fig:framework}
\end{figure}

In fact, \citet{bitar2011selling} and \citet{baeyens2013coalitional} have demonstrated that WPPs can consistently increase their expected profits through aggregated trading.
As discussed above, in the centralized context, it is optimal to submit $y_{t+k,\mathrm{sum}}^{\mathrm{F}*}$ as the trading offer, as the primary objective is to maximize total profits. In contrast, in a multi-agent setting where the generation assets are owned by different companies, each agent seeks to maximize its own profit. As a result, the aggregate of independently determined offers often diverges from the PM’s optimal offer. This discrepancy necessitates determining both the total trading offer submitted to the market and the corresponding profit allocation for each individual agent. Applying simple cost allocation rules—such as evenly splitting the total cost among WPPs—may lead to dissatisfaction among participants due to perceived unfairness. To address this issue, various cooperative game-theoretic approaches have been employed to allocate costs among participating agents, including proportional sharing, the Shapley value, and the nucleolus. For instance, \citet{freire2014hybrid, nguyen2018sharing} developed Benders decomposition and constraint-generation algorithms to solve mixed-integer linear programming formulations of the nucleolus problem, which aims to identify allocation rules within the core. Based on the selected allocation rule, the PM determines the market offer and subsequently distributes profits (or, equivalently, costs) to the WPPs. The operational framework implemented by the PM is illustrated in Figure~\ref{fig:framework}.

Although existing studies have investigated various allocation rules and computational techniques, they typically assume that the PM and WPPs have access to identical information and that WPPs unanimously agree on the submitted offers. In practice, however, diverse perspectives on future energy generation can lead to differing opinions between the PM and the WPPs regarding the appropriate offers. Specifically, each WPP typically uses a forecast model to estimate its own independent offer, as expressed in \eqref{forecast_offer}. Similarly, the PM employs a forecasting model $f$ with parameters $\phi$ to predict the aggregated offer based on the feature vector $\mathbf{x}_t$. This model is expressed as
\begin{equation}
    y^\mathrm{F}_{t+k,\mathrm{sum}}=\hat{y}_{t+k,\mathrm{sum}|t}=f(\mathbf{x}_{t};\phi).
\end{equation} 
As the forecasts $\left\{\hat{y}_{t+k,\mathrm{sum}|t},\hat{y}_{t+k,1|t},\ldots,\hat{y}_{t+k,m|t}\right\}$ are generated independently by the PM and WPPs, they are likely to be incoherent. Thus, WPPs may disagree with the offer proposed by the PM, and furthermore, the forecasts cannot be directly used to determine each WPP's share in the submitted offer. We address this challenge by developing a value-oriented forecast reconciliation approach in Section~\ref{s4}.

\section{Methodology}
\label{s4}

In this section, we first formulate the forecast reconciliation problem within the context of aggregated wind energy trading. We then propose a value-oriented forecast reconciliation approach based on the Nash bargaining framework, followed by the presentation of a primal-dual algorithm for parameter estimation.

\subsection{Problem Formulation}


For ease of reference, the following concise notations are introduced: $\left[y_{t+k,\mathrm{sum}},y_{t+k,1},\ldots,y_{t+k,m}\right]^\top$ is represented by $\mathbf{y}_{t+k}$, while $\left[\hat{y}_{t+k,\mathrm{sum}|t},\hat{y}_{t+k,1|t},\cdots,\hat{y}_{t+k,m|t}\right]^\top$ is expressed as $\hat{\mathbf{y}}_{t+k|t}$, and referred to as the \textit{base forecast}. As shown in Figure \ref{fig:framework}, the interaction between WPPs and the PM inherently establishes a two-level hierarchical structure. In the context of forecast reconciliation \citep{athanasopoulos2023forecast}, forecasts at the leaf nodes of the hierarchy are referred to as \textit{bottom-level forecasts}, while forecasts at other nodes are called \textit{aggregated forecasts}. In our specific WPP-PM hierarchy, we can decompose the vector $\mathbf{y}_t$ into $m$ bottom-level values $\mathbf{b}_t\in \mathbb{R}^m$ and an aggregated value $y_{t,\mathrm{sum}}$ such that $\mathbf{y}_t=\left[y_{t,\mathrm{sum}},\mathbf{b}_t^\top\right]^\top$. By defining a structural matrix $\mathbf{S}\in \{0,1\}^{(m+1)\times m}$, we obtain
\begin{equation}
    \mathbf{y}_t=\mathbf{S}\mathbf{b}_t \Leftrightarrow \begin{bmatrix}
y_{t,\mathrm{sum}} \\
\mathbf{b}_t 
\end{bmatrix}=\begin{bmatrix}
 \mathbf{s}_{\mathrm{sum}}\\
 \mathbf{I}_m
\end{bmatrix}\mathbf{b}_t, \ \forall t,
    \label{eq17}
\end{equation}
where $\mathbf{s}_{\mathrm{sum}}=[1,1,\cdots,1]$ is an aggregation vector and $\mathbf{I}_m$ is the $m\times m$ identity matrix. Equivalently, \eqref{eq17} can be rewritten as
\begin{equation}
    \mathbf{B}\mathbf{y}_t=\mathbf{0},
\end{equation}
where $\mathbf{B}=[1,-\mathbf{s}_{\mathrm{sum}}]$. Notably, the hierarchy does not have to be confined to just two levels, as it can accommodate $n-m>1$ aggregated series, meaning that $\mathbf{y}_t\in \mathbb{R}^n$. Correspondingly, the aggregation matrix is represented by $\mathbf{S}_{\mathrm{sum}}\in \{0,1\}^{(n-m)\times m}$, and $\mathbf{B}$ defined as $[\mathbf{I}_{n-m}|-\mathbf{S}_{\mathrm{sum}}]$.
With the hierarchical structure, the coherent forecast is formally defined as follows.
\begin{definition}[Coherent forecast]
A forecast $\hat{\mathbf{y}}_{t+k|t}$ is said to be coherent if $\mathbf{B}\hat{\mathbf{y}}_{t+k|t}=\mathbf{0}$.
\end{definition}
As previously stated, the base forecast, $\hat{\mathbf{y}}_{t+k|t}$ is independently generated by WPPs and PM, and is likely not coherent. Therefore, it is essential to reconcile the forecasts before using them in market trading. Let $\Tilde{\mathbf{y}}_{t+k|t}=\left[\Tilde{y}_{t+k,\mathrm{sum}|t},\Tilde{y}_{t+k,1|t},\ldots,\Tilde{y}_{t+k,m|t}\right]^\top$ denote the reconciled forecasts, where
\begin{equation}
    \Tilde{y}_{t+k,\mathrm{sum}|t}=\sum_{i=1}^m\Tilde{y}_{t+k,i|t}.
\end{equation}
The reconciliation process can then be conceptually expressed as
\begin{align}
    \Tilde{\mathbf{y}}_{t+k|t}=f_\mathrm{RE}(\hat{\mathbf{y}}_{t+k|t}),
\end{align}
where $f_\mathrm{RE}$ represents the reconciliation function. In other words, we transform incoherent base forecast $\hat{\mathbf{y}}_{t+k|t}$ into coherent forecast $\Tilde{\mathbf{y}}_{t+k|t}$ via the reconciliation function. This process can be viewed as a special approach to forecast combination, integrating both direct predictions for each series and indirect forecasts derived from the structure \citep{hollyman2021understanding}.

After reconciliation, the PM then places offers in the market using the reconciled aggregate forecast, $\Tilde{y}_{t+k,\mathrm{sum}|t}$. In parallel, the reconciled forecast for each WPP, $\Tilde{y}_{t+k,i|t}$, is not submitted to the market; it serves instead to determine cost (or profit) allocation and is thus termed a pseudo-offer. In contrast to the independent offering outlined in Section~\ref{s3}, where all WPPs utilize an identical cost function, namely $c_{t+k}(\hat{y}_{t+k,i},y_{t+k,i})$, in this context, the WPPs encounter distinct cost functions. This variation arises because the WPPs are allocated as portions of the total balancing cost. Concretely, the allocated cost for WPP $i$ is calculated based on the reconciled forecasts $\Tilde{\mathbf{y}}_{t+k|t}$ and realizations $\mathbf{y}_{t+k}$, which is represented as $c_{t+k,i}^{(\mathrm{AG})}(\Tilde{\mathbf{y}}_{t+k|t},\mathbf{y}_{t+k})$. The detailed definition of $c_{t+k,i}^{(\mathrm{AG})}(\Tilde{\mathbf{y}}_{t+k|t},\mathbf{y}_{t+k})$ is postponed to Section~\ref{S4.2}. Correspondingly, the profit for WPP $i$ in the context of aggregated wind energy trading, is represented as $\rho_{t+k,i}^{(\mathrm{AG})}(\Tilde{\mathbf{y}}_{t+k|t},\mathbf{y}_{t+k})$, which can be expressed as
\begin{equation}
\label{profit_agg}
    \rho_{t+k,i}^{(\mathrm{AG})}(\Tilde{\mathbf{y}}_{t+k|t},\mathbf{y}_{t+k})=\pi^\mathrm{F}_{t+k}y_{t+k,i}-c_{t+k,i}^{(\mathrm{AG})}(\Tilde{\mathbf{y}}_{t+k|t},\mathbf{y}_{t+k}).
\end{equation}

It is observed that, after reconciliation, the balancing cost of the aggregated forecast is always lower than or equal to the sum of costs of independent pseudo-offers, as described below.
\begin{proposition}
\label{theorem4.1}
Let $\left\{ \Tilde{y}_{t+k,1|t},\ldots,\Tilde{y}_{t+k,m|t}\right\}$ be a set of $m$ independent pseudo-offers for the trading hour $t+k$. Then, it holds that,
\begin{align*}
    c_{t+k}(\Tilde{y}_{t+k,\mathrm{sum}|t},y_{t+k,\mathrm{sum}})\leq\sum_{i=1}^mc_{t+k}(\Tilde{y}_{t+k,i|t},y_{t+k,i}),
\end{align*}
where $\Tilde{y}_{t+k,\mathrm{sum}|t}=\sum_{i=1}^m\Tilde{y}_{t+k,i|t}$.
\end{proposition}
\begin{proof}
See the appendix.    
\end{proof}
\noindent Since the pseudo-offers $\Tilde{y}_{t+k,i|t}$ are not actually submitted to the markets, the above proposition illustrates the potential cost implications of using reconciled forecasts as offers in the markets for any realization of the actual observations. Specifically, in the case of bottom-up reconciliation, each reconciled forecast $\Tilde{y}_{t+k,i|t}$ coincides with the corresponding base forecast $\hat{y}_{t+k,i|t}$. This implies that the balancing cost of an aggregated offering can be lower than or equal to the sum of the costs incurred under independent offerings. Based on this, we state the following corollary.
\begin{corollary}
After bottom-up reconciliation, the balancing cost of an aggregated offering is less than or equal to the sum of the costs incurred under independent offerings, for any realization of the actual observations. 
\end{corollary}

Notably, coherence is often effectively ensured using existing forecast reconciliation methods. However, as noted by \citet{athanasopoulos2023evaluation}, heterogeneous objectives sometimes arise across different levels of the hierarchy, with specific decisions tied to individual nodes. Despite this, most existing quality-oriented reconciliation approaches adopt a single loss function—typically the mean squared error—within the reconciliation process \citep{athanasopoulos2023forecast}. In contexts where agents possess heterogeneous cost functions, reconciliation naturally becomes a multi-objective optimization problem, highlighting the inherent challenge of reaching consensus on a suitable reconciliation function. Before presenting our proposed method, we outline several key assumptions, deferring detailed methodology to Section~\ref{S4.2}.

\begin{assumption}
The group of WPPs is compelled to form a grand coalition. If they fail to reach a consensus on the reconciliation process, each WPP will act independently.
\end{assumption}
\label{assum1}
\noindent In this context, we exclude sub-coalitions, as the forecasts for potential sub-coalitions are not available. Each WPP will decide whether to enter into an agency contract with the PM. Even if some WPPs choose to opt out of the contract, the PM can still manage the portfolio consisting of the remaining WPPs.
\begin{assumption}
\label{assum2}
Each WPP receives no less than the expected profit they would achieve through independent trading.
\end{assumption}
\noindent This assumption is valid because WPPs are primarily focused on the individual benefits they derive from the coalition, rather than on the collective benefits of the coalition.
As both $\alpha_{t+k}$ and $Y_{t,i}$ are time-varying, we introduce the following assumptions in line with \citet{besbes2015non}.
\begin{assumption}
\label{assum3}
We define the total variation of the sequence $\{\alpha_{t+k} \}$ as $V_\alpha$, given by
\begin{equation}
    V_\alpha=\sum_{t=1}^T|\alpha_{t+k}-\alpha_{t+k-1}|.
\end{equation}
We assume that $V_\alpha=o(T)$, implying that $\alpha_{t+k}$ varies slowly over time.
\end{assumption}
\noindent This assumption is practically reasonable, as electricity prices typically do not exhibit frequent abrupt changes.
\begin{assumption}
\label{assum4}
For each distribution sequence $\{Y_{t,i}\}$ and $\{Y_{t,\mathrm{sum}}\}$, we define the total variation distance between consecutive distributions as $V_{Y,i}$, given by
\begin{equation}
    V_{Y,i}=\sum_{t=1}^T\frac{1}{2}\int|p_{t+k,i}(y)-p_{t+k-1,i}(y)|\mathrm{d}y.
\end{equation}
We assume that $V_{Y,i}=o(T)$, implying that $Y_{t,i}$ evolves slowly over time.
\end{assumption}
\noindent This assumption permits the time series to be non-stationary, provided that the underlying process exhibits only slow drift over time. Given that wind power time series are often quasi-stationary in practice, this assumption is also well justified.

\subsection{Value-oriented Forecast Reconciliation}
\label{S4.2}

Our reconciliation model is inspired by \citet{pennings2017integrated}. Specifically, the reconciliation model $f_\mathrm{RE}$ is implemented using a structural matrix $\mathbf{S}$ combined with a learnable combination function $g$. Additionally, we incorporate contextual features, such as market data, represented as $\mathbf{z}_t\in \mathbb{R}^d$, throughout the reconciliation process. As a result, both the base forecasts $\hat{\mathbf{y}}_{t+k|t}$ and the contextual features $\mathbf{z}_t$ serve as inputs to the model $f_\mathrm{RE}$. The reconciliation procedure is characterized by
\begin{align}
\label{recon_model}
    \Tilde{\mathbf{y}}_{t+k|t} &= \mathbf{S}\mathbf{h}_{t+k},\\
    \mathbf{h}_{t+k} &= g(\hat{\mathbf{y}}_{t+k|t},\mathbf{z}_t; \boldsymbol{\theta})
\end{align}
where $\mathbf{h}_{t+k}$ denotes the latent variable and $\boldsymbol{\theta}$ symbolizes the parameters within the combination function $g$. In accordance with (\ref{eq17}), $\mathbf{h}_{t+k}$ can also be interpreted as the reconciled bottom-level forecasts. This approach generalizes the widely used reconciliation technique based on linear combinations of base forecasts \citep{hollyman2021understanding}, by allowing the input vector $\mathbf{z}_t$ to incorporate information beyond the original time series. Specifically, the learning-based reconciliation process leverages both the time series and external information to produce reconciled forecasts at the bottom level, which are then aggregated to generate forecasts at higher levels. This strategy aligns with the framework proposed by \citet{bertani2024joint}.

To estimate the parameters $\boldsymbol{\theta}$, machine learning techniques are commonly employed. This involves solving the following optimization problem, with a specific loss function denoted as $l(\cdot,\cdot)$:
\begin{equation}
    \min_{\boldsymbol{\theta}}\mathbb{E}\left[l(\Tilde{\mathbf{y}}_{t+k|t},\mathbf{y}_{t+k})\right].
\end{equation}
For example, \citet{ben2019regularized} use the widely adopted mean squared error as the loss function, aiming to minimize the total root mean square error. In contrast, here each WPP $i$ has the cost function $c_{t+k,i}^{(\mathrm{AG})}(\Tilde{\mathbf{y}}_{t+k|t},\mathbf{y}_{t+k})$ and seeks to minimize its own cost. Therefore, for a group of WPPs, they must collectively decide on the parameters $\boldsymbol{\theta}$ within the reconciliation model. In such a multi-agent system, the parameter estimation procedure, following the classical utilitarian principle, entails minimizing the total cost incurred by all WPPs, formally expressed as:
\begin{align*}
    \min_{\boldsymbol{\theta}}\sum_{i=1}^m\mathbb{E}\left[c_{t+k,i}^{(\mathrm{AG})}(\Tilde{\mathbf{y}}_{t+k|t},\mathbf{y}_{t+k})\right].
\end{align*}
The parameters obtained from this estimation are referred to as the utilitarian estimate. However, while each WPP aims to minimize its own costs, the sum of costs is neutral with respect to potential inequalities in the distribution of costs among the involved agents. Although weighting factors can be incorporated into the reconciliation process to account for forecasts at each hierarchical level \citep{van2015game}, there is a lack of well-defined methodologies for determining these weights in the context of decision-making. In contrast, incorporating fairness considerations into the parameter estimation process offers a viable and rigorous alternative \citep{bertsimas2011price}. To achieve this, we use the cooperative bargaining framework to distribute costs fairly among the agents involved.

\subsubsection{Nash Bargaining based Reconciliation}

In a bargaining game of $m$ agents, the agreement set $\mathcal{A}$ is a set of all agreed parameters $\boldsymbol{\theta}$, whereas the disagreement point $\mathbf{d}$ is the default decision when the group of agents cannot reach a consensus.
Therefore, we redefine the loss function for each WPP $i$ as
\begin{equation}
l_i(\boldsymbol{\theta})=\left\{
\begin{aligned}
\mathbb{E}\left[c_{t+k,i}^{(\mathrm{AG})}(f_\mathrm{RE}(\hat{\mathbf{y}}_{t+k|t},\mathbf{z}_t;\boldsymbol{\theta}),\mathbf{y}_{t+k})\right],& \ \mathrm{if} \ \boldsymbol{\theta} \in \mathcal{A}, \\
\mathbb{E}\left[c_{t+k,i}(\hat{y}_{t+k,i},y_{t+k,i})\right],& \ \mathrm{if} \ \boldsymbol{\theta}=\mathbf{d}.
\end{aligned}
\right.
\end{equation}
This implies that when the participating agents reach a consensus, they reconcile their forecasts and engage in collective market trading, thereby sharing the associated costs. Under Assumption \ref{assum1}, sub-coalitions are not permitted; thus, in the absence of consensus, each agent trades independently. The disagreement point $\mathbf{d}$ can be assigned an unattainable value, such as an all-zero vector, since the reconciliation of forecasts is no longer required. Specifically, we assume that within the agreement set, there are points that present Pareto improvements\footnote{Consider a set of objective functions $l_1,l_2,\cdots,l_m$ to be minimized with respect to a parameter vector $\boldsymbol{\theta}$. We say that $\boldsymbol{\theta}$ is a Pareto improvement over $\boldsymbol{\theta}^\prime\in \mathcal{A}$, if $l_i(\boldsymbol{\theta}) \leq l_i(\boldsymbol{\theta}^\prime) ,\forall i=1,2\cdots,m$, and the inequality is strict for at least one index $i$.} over the disagreement point. If no such points exist, individual rationality will be violated, thereby rendering the design of a reconciliation function meaningless. This assumption is intuitively plausible, as Theorem \ref{theorem3.1} demonstrates how aggregate trading can reduce overall balancing costs. We will show that it holds within the framework of our allocation rule under a few mild conditions.

Particularly, the Nash bargaining framework proposes four axioms, namely Pareto optimality, symmetry, independence of irrelevant alternatives, and invariance to affine transformations \citep{nash1953two}, which are described as follows:
\begin{enumerate}[(1)]
    \item Pareto optimality: the solution $\boldsymbol{\theta}^\mathrm{NB}$ is Pareto optimal in $\mathcal{A}$, that is, there does not exist $\boldsymbol{\theta}\in \mathcal{A}$ such that $l_i(\boldsymbol{\theta})<l_i(\boldsymbol{\theta}^\mathrm{NB}),\ \forall i=1,2,\cdots,m$, and $\boldsymbol{\theta} \neq \boldsymbol{\theta}^\mathrm{NB}$.
    \item Symmetry: the solution $\boldsymbol{\theta}^\mathrm{NB}$ is invariant in the order of agents.
    \item Independence of irrelevant alternatives: if we expand the feasible set $\mathcal{A}$ and the solution stays in $\mathcal{A}$, then the solution $\boldsymbol{\theta}^\mathrm{NB}$ stays the same.
    \item Invariance to affine transformations: if affine transformations are applied to the costs, the fair allocation in the affine-transformed system equates to the affine transformation of the fair allocation in the original system.
\end{enumerate}

According to \citet{nash1953two}, a unique solution exists for the bargaining game. In the context of the cost sharing game defined in this study, the Nash bargaining solution can be derived as follows:
\begin{subequations}
\label{Nb_framework}
\begin{align}
    \boldsymbol{\theta}^\mathrm{NB}=\arg \underset{\boldsymbol{\theta}\in \mathcal{A}}{\min} \ &\sum_i^m -\log\left[l_i(\mathbf{d})-l_i(\boldsymbol{\theta}) \right] \\
    \mathrm{s.t.} \ & l_i(\mathbf{d}) \geq l_i(\boldsymbol{\theta}), \ i=1,2,\cdots,m.
\end{align}
\end{subequations}
The constraint in (\ref{Nb_framework}) ensures that the allocated cost for each agent does not exceed the cost incurred under independent offering, as stipulated by Assumption \ref{assum2}.

In fact, neither $\mathbb{E}\left[c_{t+k,i}^{(\mathrm{AG})}(f_\mathrm{RE}(\hat{\mathbf{y}}_{t+k|t},\mathbf{z}_t;\boldsymbol{\theta}),\mathbf{y}_{t+k})\right]$ nor $\mathbb{E}\left[c_{t+k}(\hat{y}_{t+k,i},y_{t+k,i})\right]$ is available. Following \citet{ban2019big}, we employ the sample average approximation method to approximate the expectation components in (\ref{Nb_framework}) and apply the empirical risk minimization algorithm for parameter estimation. As a result, we convert problem (\ref{Nb_framework}) into a constrained learning problem. With a set of historical data $\{\hat{\mathbf{y}}_{t+k|t},\mathbf{z}_t,\mathbf{y}_{t+k} \}_{t=1,2,\cdots,T}$, we can rewrite (\ref{Nb_framework}) as
\begin{subequations}
\label{erm-reconciliation} 
\begin{align} 
    \underset{\boldsymbol{\theta}}{\min} & \, \quad \sum_i^m -\log\left[\frac{1}{T}\sum_{t=1}^T\left[c_{t+k}(\hat{y}_{t+k,i|t},y_{t+k,i})-c_{t+k,i}^{(\mathrm{AG})}(\Tilde{\mathbf{y}}_{t+k|t},\mathbf{y}_{t+k})\right] \right], \label{erm_obj} & \\
  \quad  \mathrm{s.t. } & \quad \, \frac{1}{T}\sum_{t=1}^T\left[c_{t+k,i}^{(\mathrm{AG})}(\Tilde{\mathbf{y}}_{t+k|t},\mathbf{y}_{t+k})-c_{t+k}(\hat{y}_{t+k,i},y_{t+k,i})\right] \leq 0, \ \forall i=1,\cdots,m, \\
  & \quad \ \Tilde{\mathbf{y}}_{t+k|t}=\mathbf{S}g(\hat{\mathbf{y}}_{t+k|t},\mathbf{z}_t;\boldsymbol{\theta}).
\end{align}
\end{subequations}
Unlike \citet{ban2019big}, which employs linear models for data-driven decisions, we leverage neural networks to model the combination function $g$, as described in (\ref{recon_model}). In the following subsections, we describe the allocation rule used and defer the details of model training to Section~\ref{S3.3}.

\subsubsection{Weighted Proportional Allocation Mechanism}
\label{s4.2.2}

We design the cost allocation rule $c_{t+k,i}^{(\mathrm{AG})}(\Tilde{\mathbf{y}}_{t+k|t},\mathbf{y}_{t+k})$ for each WPP within the framework of value-oriented forecast reconciliation. Specifically, we employ the weighted proportional allocation rule introduced by \citet{anupindi2001general}. This rule consists of two components: one associated with the bottom-level reconciled forecast $\Tilde{y}_{t+k,i|t}$ and the other with the aggregate-level reconciled forecast $\Tilde{y}_{t+k,\mathrm{sum}|t}$. The cost $c_{t+k,i}^{(\mathrm{AG})}(\Tilde{\mathbf{y}}_{t+k|t},\mathbf{y}_{t+k})$ is mathematically expressed as
\begin{equation}
\label{allocation}
    c_{t+k,i}^{(\mathrm{AG})}(\Tilde{\mathbf{y}}_{t+k|t},\mathbf{y}_{t+k}) \, = \, (1-w) \, c_{t+k}(\Tilde{y}_{t+k,i|t},y_{t+k,i})\, +\, w \, \gamma_i \, c_{t+k}(\Tilde{y}_{t+k,\mathrm{sum}|t},y_{t+k,\mathrm{sum}}).
\end{equation}
where $0\leq w \leq 1$, $\gamma_i \geq0$, and $\sum_i^m\gamma_i=1$. The term $c_{t+k}(\Tilde{y}_{t+k,i|t},y_{t+k,i})$ represents the cost associated with the WPP $i$ if it independently submits the pseudo-offer $\Tilde{y}_{t+k,i|t}$ in the markets. In contrast, $c_{t+k}(\Tilde{y}_{t+k,\mathrm{sum}|t},y_{t+k,\mathrm{sum}})$ represents the total cost for the aggregation when the PM submits the offer $\Tilde{y}_{t+k,\mathrm{sum}|t}$ in the markets.

This allocation rule implies that each WPP reduces a fraction of its own cost associated with the pseudo-offer while simultaneously bearing an equivalent fraction of its share of the overall system cost, $\gamma_ic_{t+k}(\Tilde{y}_{t+k,\mathrm{sum}|t},y_{t+k,\mathrm{sum}})$. When $w=0$, the PM retains the entire profit from the aggregate trading and distributes none to the WPPs. Consequently, each WPP acts independently using the reconciled forecast $\Tilde{y}_{t+k,i|t}$, as if aggregation does not exist. In contrast, when $w=1$, the cost incurred by each WPP is represented by $\gamma_ic_{t+k}(\Tilde{y}_{t+k,\mathrm{sum}|t},y_{t+k,\mathrm{sum}})$, which denotes a portion of the system-wide cost. For $w\in (0,1)$, the weighted proportional allocation rule reflects allocation schemes that fall between these extreme cases.

Specifically, $\gamma_i$ can be defined according to various criteria, as outlined by \citet{hartman1996cost}. For example, one definition uses the proportion of actual power generation\footnote{When $y_{t+k,\mathrm{sum}}$ is equal to 0, we set $\gamma_i^\mathrm{ge}$ as $1/m$.}, given by:
\begin{align*}
    \gamma_i^\mathrm{ge}=\frac{y_{t+k,i}}{y_{t+k,\mathrm{sum}}}.
\end{align*}
Another definition considers $\gamma_i$ as the proportion of cost associated with pseudo offers, expressed as:
\begin{align*}
    \gamma_i^\mathrm{pc}=\frac{c_{t+k}(\Tilde{y}_{t+k,i|t},y_{t+k,i})}{\sum_{i=1}^mc_{t+k}(\Tilde{y}_{t+k,i|t},y_{t+k,i})}.
\end{align*}
It is observed that these settings of $\gamma_i$ ensure the allocated cost for WPP $i$ is less than its corresponding cost associated with pseudo offers, as stated below.
\begin{proposition}
\label{P4.1}
Designate $\gamma_i$ in the cost allocation formula (\ref{allocation}) as $\gamma_i^\mathrm{pc}$. Thus, we obtain the following inequality:
\begin{align*}
    c_{t+k,i}^{(\mathrm{AG})}(\Tilde{\mathbf{y}}_{t+k|t},\mathbf{y}_{t+k})\leq c_{t+k}(\Tilde{y}_{t+k,i|t},y_{t+k,i}),
\end{align*}
which ensures that the cost allocation for WPP $i$ is less than or equal to the corresponding cost associated with its pseudo offers.
\end{proposition}
\begin{proof}
See the appendix.    
\end{proof}
\begin{proposition}
\label{P4.2}
Define $\gamma_i$ in the cost allocation (\ref{allocation}) as $\gamma_i^\mathrm{ge}$. If the following condition holds: 
\begin{align}
\label{unit_cost cond}
    \frac{c_{t+k}(\Tilde{y}_{t+k,\mathrm{sum}|t},y_{t+k,\mathrm{sum}})}{y_{t+k,\mathrm{sum}}}\leq \frac{c_{t+k}(\Tilde{y}_{t+k,i|t},y_{t+k,i})}{y_{t+k,i}}, \ \forall i=1,2,\ldots,m,
\end{align}
then it follows that
\begin{align*}
    c_{t+k,i}^{(\mathrm{AG})}(\Tilde{\mathbf{y}}_{t+k|t},\mathbf{y}_{t+k})\leq c_{t+k}(\Tilde{y}_{t+k,i|t},y_{t+k,i}).
\end{align*}
\end{proposition}
\noindent The proof is straightforward and thus omitted here. In practice, the condition (\ref{unit_cost cond}) is typically satisfied, as demonstrated through simulations and case studies. If the cost $c_{t+k}(\Tilde{y}_{t+k,i|t},y_{t+k,i})$ for the pseudo offer $\Tilde{y}_{t+k,i|t}$ is lower than that of the independent offer $c_{t+k}(\hat{y}_{t+k,i|t},y_{t+k,i})$, then the aforementioned propositions suggest that the cost experienced by WPP $i$ under the proposed method, represented by $c_{t+k,i}^{(\mathrm{AG})}(\Tilde{\mathbf{y}}_{t+k|t},\mathbf{y}_{t+k})$, is less than $c_{t+k}(\hat{y}_{t+k,i|t},y_{t+k,i})$. Nevertheless, it implies a reduction in the cost of independent offers post-reconciliation, which is a strong requirement. 

We note that Propositions \ref{P4.1} and \ref{P4.2} hold for any parameters $\boldsymbol{\theta}$. Particularly, in the context of the BU reconciliation approach, the bottom-level forecasts remain unchanged after reconciliation, i.e., $\Tilde{y}_{t+k,i|t}=\hat{y}_{t+k,i|t}$. Consequently, it follows that the shared allocation cost $c_{t+k,i}^{(\mathrm{AG})}(\Tilde{\mathbf{y}}_{t+k|t},\mathbf{y}_{t+k})$ is less than or equal to the cost of independent offering $c_{t+k}(\hat{y}_{t+k,i|t},y_{t+k,i})$. It is evident that,
\begin{align*}
    \mathbb{E}\left[c_{t+k,i}^{(\mathrm{AG})}(\Tilde{\mathbf{y}}_{t+k|t},\mathbf{y}_{t+k})\right]\leq \mathbb{E}\left[c_{t+k}(\hat{y}_{t+k,i|t},y_{t+k,i})\right].
\end{align*} 
Based on this result, we derive the following corollary.
\begin{corollary}
\label{corollary2}
By employing the weighted proportional allocation rule, the agreement set of parameters $\mathbf{\theta}$ in problem (\ref{Nb_framework}) is guaranteed to be non-empty.
\end{corollary}

\noindent Given that both $c_{t+k}(\Tilde{y}_{t+k,i|t},y_{t+k,i})$ and $c_{t+k}(\Tilde{y}_{t+k,\mathrm{sum}|t},y_{t+k,\mathrm{sum}})$ are convex functions, introducing $\gamma_i^\mathrm{pc}$ disrupts the convexity of the cost function $c_{t+k,i}^{(\mathrm{AG})}(\Tilde{\mathbf{y}}_{t+k|t},\mathbf{y}_{t+k})$, as the ratio of convex functions is generally not convex. In contrast, $\gamma_i^\mathrm{ge}$ is a constant and thus preserves the convexity of $c_{t+k,i}^{(\mathrm{AG})}(\Tilde{\mathbf{y}}_{t+k|t},\mathbf{y}_{t+k})$. For this reason, we primarily adopt $\gamma_i^\mathrm{ge}$ throughout this work.

\subsubsection{Profit Allocation within the Aggregation}
To further analyze the profit allocations, we formally define the extra profit from the forecast reconciliation as follows.
\begin{definition}
[Extra profit]    
Let $\Tilde{y}_{t+k,\mathrm{sum}|t}$ be the offer by the PM for the trading hour $t+k$, and ${\hat{y}_{t+k,1|t},\cdots,\hat{y}_{t+k,m|t}}$ be a set of $m$ independent offers. Then, the extra profit is defined as
\begin{align*}
    R=\sum_{i=1}^mc_{t+k}(\hat{y}_{t+k,i|t},y_{t+k,i})-c_{t+k}(\Tilde{y}_{t+k,\mathrm{sum}|t},y_{t+k,\mathrm{sum}}),
\end{align*}
where $c_{t+k}(\Tilde{y}_{t+k,\mathrm{sum}|t},y_{t+k,\mathrm{sum}})$ is the cost for the aggregate offer $\Tilde{y}_{t+k,\mathrm{sum}|t}$ post-reconciliation, and $c_{t+k}(\hat{y}_{t+k,i|t},y_{t+k,i})$ is the cost for the independent offer using the  base forecast $\hat{y}_{t+k,i|t}$.
\end{definition}
\noindent This definition calculates the difference between the total costs of independent offers without reconciliation (i.e., disagreement point) and the cost of the aggregate offer by the PM post-reconciliation (i.e., agreement point). 
In parallel, we calculate the total cost assigned to all WPPs as
\begin{align*}
    \sum_{i=1}^mc_{t+k,i}^{(\mathrm{AG})}(\Tilde{\mathbf{y}}_{t+k|t},\mathbf{y}_{t+k})=(1-w)\sum_{i=1}^mc_{t+k}(\Tilde{y}_{t+k,i|t},y_{t+k,i})+wc_{t+k}(\Tilde{y}_{t+k,\mathrm{sum}|t},y_{t+k,\mathrm{sum}}).
\end{align*}
It is noteworthy that the total cost of WPPs is higher than the actual balancing cost incurred by the PM, but remains lower than the sum of the cost of the pseudo offers, i.e.,
\begin{align}
    c_{t+k}(\Tilde{y}_{t+k,\mathrm{sum}|t},y_{t+k,\mathrm{sum}})\leq \sum_{i=1}^mc_{t+k,i}^{(\mathrm{AG})}(\Tilde{\mathbf{y}}_{t+k|t},\mathbf{y}_{t+k})\leq \sum_{i=1}^mc_{t+k}(\Tilde{y}_{t+k,i|t},y_{t+k,i}),
\end{align}
where the inequality results from Proposition \ref{theorem4.1}. Therefore, it can be seen that the designed cost allocation rule is not efficient, unless $w=1$, as we have
\begin{equation}
\sum_{i=1}^mc_{t+k}(\hat{y}_{t+k,i|t},y_{t+k,i})-\sum_{i=1}^mc_{t+k,i}^{(\mathrm{AG})}(\Tilde{\mathbf{y}}_{t+k|t},\mathbf{y}_{t+k})\leq R.
\end{equation}
In other words, the total extra profit allocated to WPPs is lower than the extra profit generated by aggregated trading. Specifically, the remaining extra profit is assigned to the PM as the payoff for their coordination efforts, denoted as $R^\mathrm{pm}$, i.e.,
\begin{align}
    R^\mathrm{pm}=(1-w)\left[\sum_{i=1}^mc_{t+k}(\Tilde{y}_{t+k,i|t},y_{t+k,i})-c_{t+k}(\Tilde{y}_{t+k,\mathrm{sum}|t},y_{t+k,\mathrm{sum}})\right].
\end{align}


\subsection{Model Training via Empirical Risk Minimization}
\label{S3.3}

Indeed, problem (\ref{erm-reconciliation}) can be viewed as a machine learning problem with constraints. As demonstrated by \citet{chamon2020probably}, learning with constraints is, in theory, no more difficult than unconstrained learning, and empirical risk minimization remains an appropriate approach for such problems. Specifically, they proposed a primal–dual gradient algorithm for constrained learning. Building on this, we adapt their framework and derive the empirical Lagrangian function for problem~(\ref{erm-reconciliation}), i.e.
\begin{equation}
\begin{split}
L(\boldsymbol{\theta},\boldsymbol{\mu})&=-\sum_i^m \log\left[\frac{1}{T}\sum_{t=1}^T\left[c_{t+k}(\hat{y}_{t+k,i|t},y_{t+k,i})-c_{t+k,i}^{(\mathrm{AG})}(\Tilde{\mathbf{y}}_{t+k|t},\mathbf{y}_{t+k})\right] \right]\\
&+\sum_i^m\mu_i\left[\frac{1}{T}\sum_{t=1}^T\left[c_{t+k,i}^{(\mathrm{AG})}(\Tilde{\mathbf{y}}_{t+k|t},\mathbf{y}_{t+k})-c_{t+k}(\hat{y}_{t+k,i|t},y_{t+k,i})\right] \right]^+,
\end{split}
\end{equation}
where $\boldsymbol{\mu}=[\mu_1,\mu_2,\cdots,\mu_m]^\top\in\mathbb{R}^m$ collects the dual variables $\mu_i$ associated with constraints. The empirical dual problem is then written as
\begin{equation}
\label{dual}
    \underset{\boldsymbol{\mu}}{\max}\ \underset{\boldsymbol{\theta}}{\min}\ L(\boldsymbol{\theta},\boldsymbol{\mu}).
\end{equation}

Obviously, the outer maximization is a convex optimization problem, and its gradient can be easily obtained by evaluating the minimizer of $L(\boldsymbol{\theta},\boldsymbol{\mu})$. For instance, given the $\boldsymbol{\theta}$, the gradient of $\mu_i$ is calculated as
\begin{align*}
    \nabla_{\mu_i}=\left[\frac{1}{T}\sum_{t=1}^T\left[c_{t+k,i}^{(\mathrm{AG})}(\Tilde{\mathbf{y}}_{t+k|t},\mathbf{y}_{t+k})-c_{t+k}(\hat{y}_{t+k,i|t},y_{t+k,i})\right] \right]^+,
\end{align*}
where $\Tilde{\mathbf{y}}_{t+k|t}=\mathbf{S}g(\hat{\mathbf{y}}_{t+k|t},\mathbf{z}_t;\boldsymbol{\theta})$.
Hence, the main challenge of problem (\ref{dual}) lies in the inner minimization. However, abundant evidence has shown that the gradient descent algorithm can learn good parameters for neural networks \citep{shalev2014understanding}. Subsequently, we can iteratively minimize the Lagrangian with respect to $\boldsymbol{\theta}$ while keeping $\boldsymbol{\mu}$ constant, followed by adjusting the dual variables using the obtained minimizer.

\begin{algorithm}[h]
\caption{Primal-dual based iterative learning algorithm}\label{alg1}
\begin{algorithmic}
\State Input: Optimizer and Lagrangian step sizes, $\lambda$, $(\nu_1,\nu_2,\cdots,\nu_m)$
\State Initialize variables: $\mu_i^0\leftarrow 1,\forall i=1,2\cdots,m$
\For{epoch $\eta=1,2,\dots$}
    \State Randomly sample a batch of samples $\{\hat{\mathbf{y}}_{t+k|t},\mathbf{z}_t,\mathbf{y}_{t+k} \}_{t\in \mathcal{T}_b}$ 
    \State    $\Tilde{\mathbf{y}}_{t+k|t}=\mathbf{S}g(\hat{\mathbf{y}}_{t+k|t},\mathbf{z}_t;\boldsymbol{\theta}^{\eta-1}),\forall t\in \mathcal{T}_b$
    \State $\boldsymbol{\theta}^\eta\leftarrow\boldsymbol{\theta}^{\eta-1}-\lambda\nabla_{\boldsymbol{\theta}} L^{(b)}(\boldsymbol{\theta},\boldsymbol{\mu}^{\eta-1})$
    \State
    $\mu_i^\eta\leftarrow\mu_i^{\eta-1}+\nu_i\left[\frac{1}{|\mathcal{T}_b|}\sum_{t\in \mathcal{T}_b}\left[c_{t+k,i}^{(\mathrm{AG})}(\Tilde{\mathbf{y}}_{t+k|t},\mathbf{y}_{t+k})-c_{t+k}(\hat{y}_{t+k,i|t},y_{t+k,i})\right] \right]^+$
\EndFor
\end{algorithmic}
\end{algorithm}

We now analyze the convexity of the Lagrangian function. The convexity of the second component of $L(\boldsymbol{\theta},\boldsymbol{\mu})$ is straightforward, as $c_{t+k,i}^{(\mathrm{AG})}(\Tilde{\mathbf{y}}_{t+k|t},\mathbf{y}_{t+k})$ is convex in $\Tilde{\mathbf{y}}_{t+k|t}$. For the first component of $L(\boldsymbol{\theta},\boldsymbol{\mu})$, we note that $-\log(\cdot)$ is convex and decreasing, while $-c_{t+k,i}^{(\mathrm{AG})}(\Tilde{\mathbf{y}}_{t+k|t},\mathbf{y}_{t+k})$ is concave in $\Tilde{\mathbf{y}}_{t+k|t}$. Since the combination of a convex, non-increasing function with a concave function preserves convexity \citep{boyd2004convex}, it follows that the first term in $L(\boldsymbol{\theta},\boldsymbol{\mu})$ is also convex. Therefore, the Lagrangian function $L(\boldsymbol{\theta},\boldsymbol{\mu})$ is convex in $\Tilde{\mathbf{y}}_{t+k|t}$, enabling the use of the stochastic gradient descent algorithm to update $\boldsymbol{\theta}$ \citep{shalev2014understanding}. The explicit form of $L^{(b)}(\boldsymbol{\theta},\boldsymbol{\mu})$ is deferred to the Appendix. Letting $\boldsymbol{\mu}^\eta$ and $\boldsymbol{\theta}^\eta$ denote the parameter estimates at the $\eta$ iteration, the iterative learning procedure is summarized in Algorithm~\ref{alg1}.

\begin{remark}
Note that the loss function is non-stationary. However, under Assumptions \ref{assum3} and \ref{assum4}, Algorithm~\ref{alg1} is guaranteed to converge sublinearly, as established by Theorem 1 by \citet{besbes2015non}.    
\end{remark}

\section{Simulations}
\label{s5}
In this section, we present several illustrative examples to validate the proposed approach using data from the Wind Toolkit\footnote{https://www.nrel.gov/grid/wind-toolkit.html}. Specifically, we have collected hourly datasets from four nearby wind farms, with capacities of 1.7496 MW, 2.9646 MW, 3.3777 MW, and 2.5272 MW, covering the period from 2007 to 2008. In particular, we follow the framework of \citet{munoz2023online}, which examines an hourly online forward market followed by a balancing market with a dual-price settlement mechanism. The lead time, $k$, for both the base forecast and forecast reconciliation process is set to 1. The dataset is split into two segments: the first 80\% is used for model training, while the remaining 20\% is reserved for out-of-sample forecasting validation. In the following, we will introduce the setup and present the simulation results.


\subsection{Experimental Setups}
In our simulation studies, we investigate a simplified scenario where $\pi_{t+k}^\mathrm{F}$ is fixed at 25 \euro/MWh, $\psi^+_{t+k}$ is set to 12 \euro/MWh, and $\psi^-_{t+k}$ is maintained at 4 \euro/MWh. This simplified setup is used to gain insight into the behavior of the proposed method, with a more practical case study deferred to Section \ref{s6}. In this context, the optimal offering strategy is to use the 0.75 quantile of the wind power distributions. Specifically, this represents the independent optimal offering strategy for each WPP and the PM. Therefore, optimal offers can be constructed using quantiles, which represent a specific type of point forecast \citep{gneiting2011quantiles}. Moreover, as point forecasts derived from mean regression are still widely used in practice, we consider both means and quantiles as base forecasts in our framework. While WPPs may act strategically by explicitly communicating quantiles, we assume that the PM remains agnostic to the specific form of the base forecasts. A description of the cases is provided below.
\begin{description}
    \item[Case 1:] Forecasts at both the aggregate and leaf levels are communicated as means, derived from mean regression models.
    \item[Case 2:] Base forecasts for both the aggregate and leaf levels are communicated as quantiles, derived from quantile regression models.\footnote{Bottom-up and quality-oriented reconciliation methods are behaviorally plausible but decision-theoretically sub-optimal in this case, which is the tension we intend to highlight.}
\end{description}

In this study, we evaluate various offering strategies, which include both independent and aggregate offering using reconciliation. The offering strategies under consideration are described as follows: \textbf{(1) Independent offering}: Each WPP submits offers based on their individual forecasts (i.e., the base forecasts). Thus, neither reconciliation nor the involvement of a PM is considered. We also include 3 aggregate offering strategies, in which the reconciled aggregate forecasts are used by the PM for offering. Concretely, the considered reconciliation approaches are described as follows. \textbf{(2) Bottom-up reconciliation}: The base forecasts at the leaf level are retained unchanged and then aggregated to generate the overall forecast. \textbf{(3) Quality-oriented reconciliation}: The base forecasts are reconciled within a learning-based model (\ref{recon_model}), where the loss function is defined as the mean squared error. \textbf{(4) Value-oriented reconciliation}: The base forecasts are reconciled using the learning-based model (\ref{recon_model}), with the training procedure outlined in Section \ref{s4}. In addition, we establish linear counterparts for both quality-oriented and value-oriented reconciliation in Case 1, where the quality-oriented model reduces to the framework proposed by \citet{hyndman2011optimal}.



It is important to note that all of the previously discussed aggregate offering strategies employ the proportional allocation rule and set $\gamma_i$ as $\gamma_i^\mathrm{ge}$, as outlined in Section \ref{S4.2}. For both quality-oriented and value-oriented reconciliation models, we include the lagged wind power generation at both the leaf and aggregate levels as contextual features.
To evaluate the performance of the offering strategies across a set of samples $\{(\hat{\mathbf{y}}_{t+k|t},\mathbf{z}_t,\mathbf{y}_{t+k})  \}_{t \in \mathcal{T}_\mathrm{test}}$, we compute the average profit (AP) for each WPP over the test set. Specifically, for independent offering, using $\rho_{t+k}$ defined by (\ref{profit_ind_2}), the average profit for WPP $i$ is given by
\begin{align*}
    \mathrm{AP}_i=\frac{1}{|\mathcal{T}_\mathrm{test}|}\sum_{t\in\mathcal{T}_\mathrm{test}}\rho_{t+k}(\hat{y}_{t+k,i|t},y_{t+k,i}).
\end{align*}
In contrast, for portfolio-based aggregate offering, using $\rho_{t+k,i}^{(\mathrm{AG})}$ from (\ref{profit_agg}), the average profit for WPP $i$ is given by
\begin{align*}
    \mathrm{AP}_i=\frac{1}{|\mathcal{T}_\mathrm{test}|}\sum_{t\in\mathcal{T}_\mathrm{test}}\rho_{t+k,i}^{(\mathrm{AG})}(\Tilde{\mathbf{y}}_{t+k|t},\mathbf{y}_{t+k}).
\end{align*}

\subsection{Case 1}

\begin{figure}[h]
    \centering
    \includegraphics[width=0.9\linewidth]{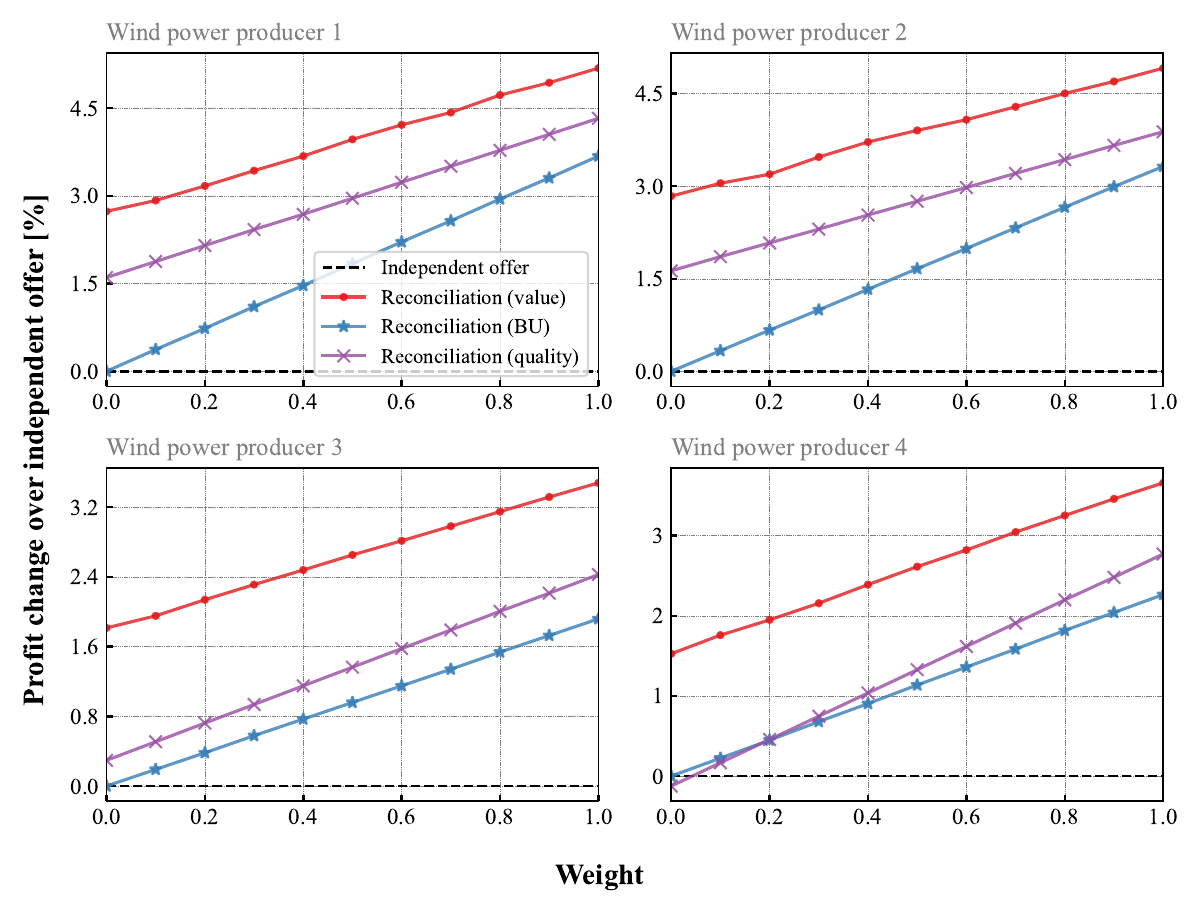}
    \caption{Average profit change over independent offers for each wind power producer across different models in Case 1}
    \label{fig:case 1 av_profit}
\end{figure}

We investigate the range of weights in the allocation rule, which varies between 0 and 1, and present the average profit over independent offers for each WPP across various models in Figure \ref{fig:case 1 av_profit}. Overall, each WPP’s profit under independent offerings is invariant to $w$, whereas profits under reconciliation-based strategies increase as $w$ rises. We have confirmed that the unit balancing cost condition (\ref{unit_cost cond}) holds in this simulation. As expected by Proposition \ref{P4.2}, the function $c_{t+k,i}^{(\mathrm{AG})}(\Tilde{\mathbf{y}}_{t+k|t},\mathbf{y}_{t+k})$ decreases monotonically with increasing $w$, which explains the observed monotonic increase in profits with respect to $w$. Specifically, within the frameworks of both BU reconciliation and quality-oriented reconciliation, profits are directly proportional to $w$. Notably, the quality-oriented reconciliation training is independent of $w$, and similarly, BU reconciliation does not require training. Thus, the balancing cost terms in (\ref{allocation}) remain unaffected by changes in $w$, leading to a linear relationship.

\begin{figure}[h]
    \centering
    \includegraphics[width=0.9\linewidth]{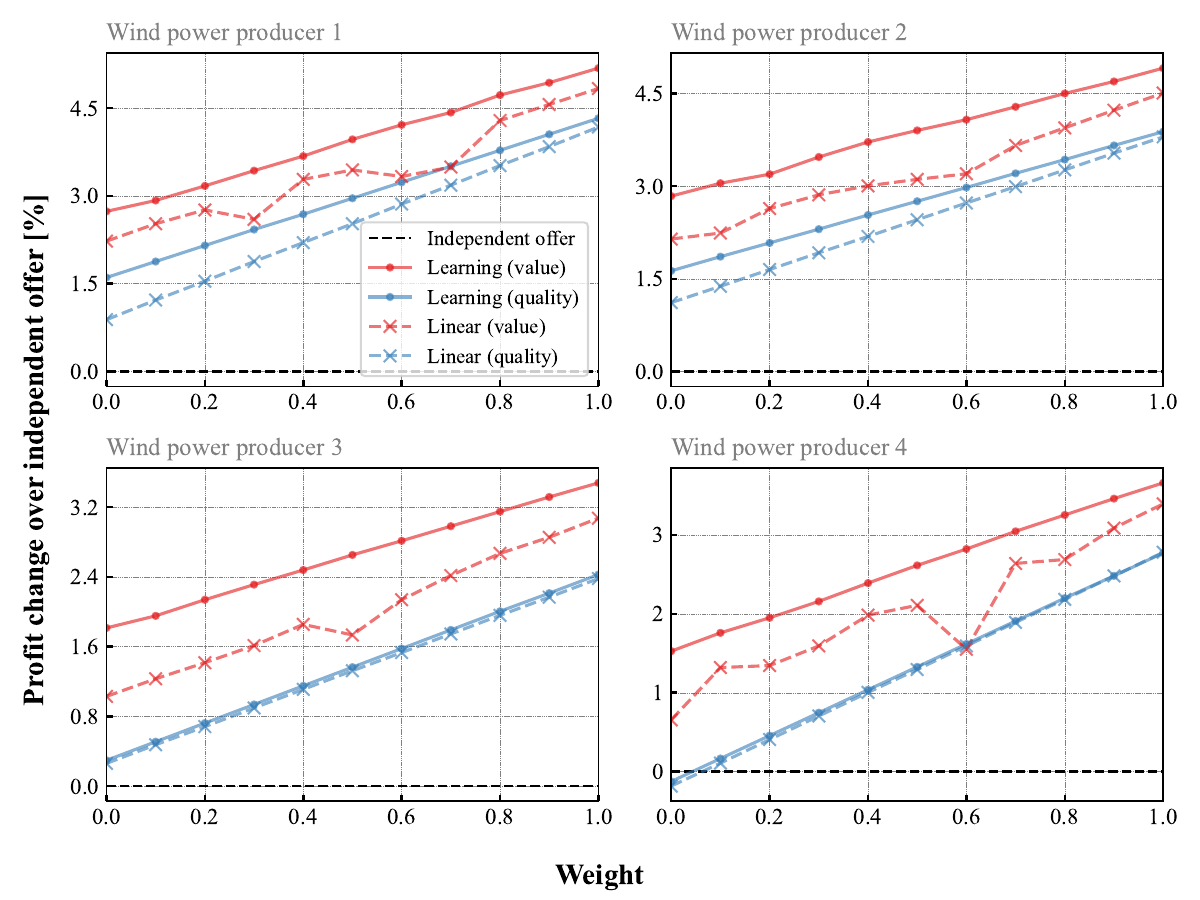}
    \caption{Average profit change over independent offers for each wind power producer under learning-based and linear reconciliation in Case 1}
    \label{fig:case 1 av_profit_com}
\end{figure}

Additionally, nearly every reconciliation-based model results in higher profits for all WPPs, except in the case of quality-focused forecast reconciliation when $w$ equals $0$ or $0.1$. Given that the reconciled forecasts at the bottom level remain unchanged in the BU reconciliation framework, it can be inferred that the profits exceed those from independent offers when condition (\ref{unit_cost cond}) is satisfied. Comparing the profits from BU reconciliation to those from quality-oriented reconciliation reveals that quality-based reconciliation generally yields higher profits, except for WPP $4$. This outcome is attributed to the ability of quality-oriented reconciliation to reduce the overall mean squared errors more effectively than BU reconciliation, as supported by existing research \citep{van2015game}. However, the profits for WPP $4$ suggest that quality-oriented reconciliation might lead to an unfair allocation of costs. Specifically, the value-oriented forecast reconciliation generates greater profits compared to other models because its loss function in the reconciliation process aligns with balancing costs. Evidently, the gains from aggregated trading differ across WPPs. In particular, WPPs 1 and 2 realize larger percentage profit increases relative to their independent offerings. This is consistent with the Nash product objective, which equalizes and safeguards excess profits over the independent baseline but does not enforce identical percentage gains. Because independent-offer profits differ across WPPs, even identical absolute excess profits translate into different percentage changes.

We present the results under both learning-based and linear reconciliation methods in Figure~\ref{fig:case 1 av_profit_com}. Overall, value-oriented reconciliation yields higher profits for each WPP compared to quality-oriented reconciliation, supporting the claim that value-oriented reconciliation is agnostic to the choice of reconciliation model. Notably, the performance of quality-oriented reconciliation is similar across both learning-based and linear implementations, highlighting the effectiveness of the optimal reconciliation approach proposed by \citet{hyndman2011optimal}. In contrast, for value-oriented reconciliation, the learning-based model outperforms its linear counterpart in both performance and stability. This is expected, as value-oriented reconciliation involves a more complex training procedure, and neural networks offer greater flexibility in function approximation.

\begin{figure}[h]
    \centering
    \includegraphics[width=0.9\linewidth]{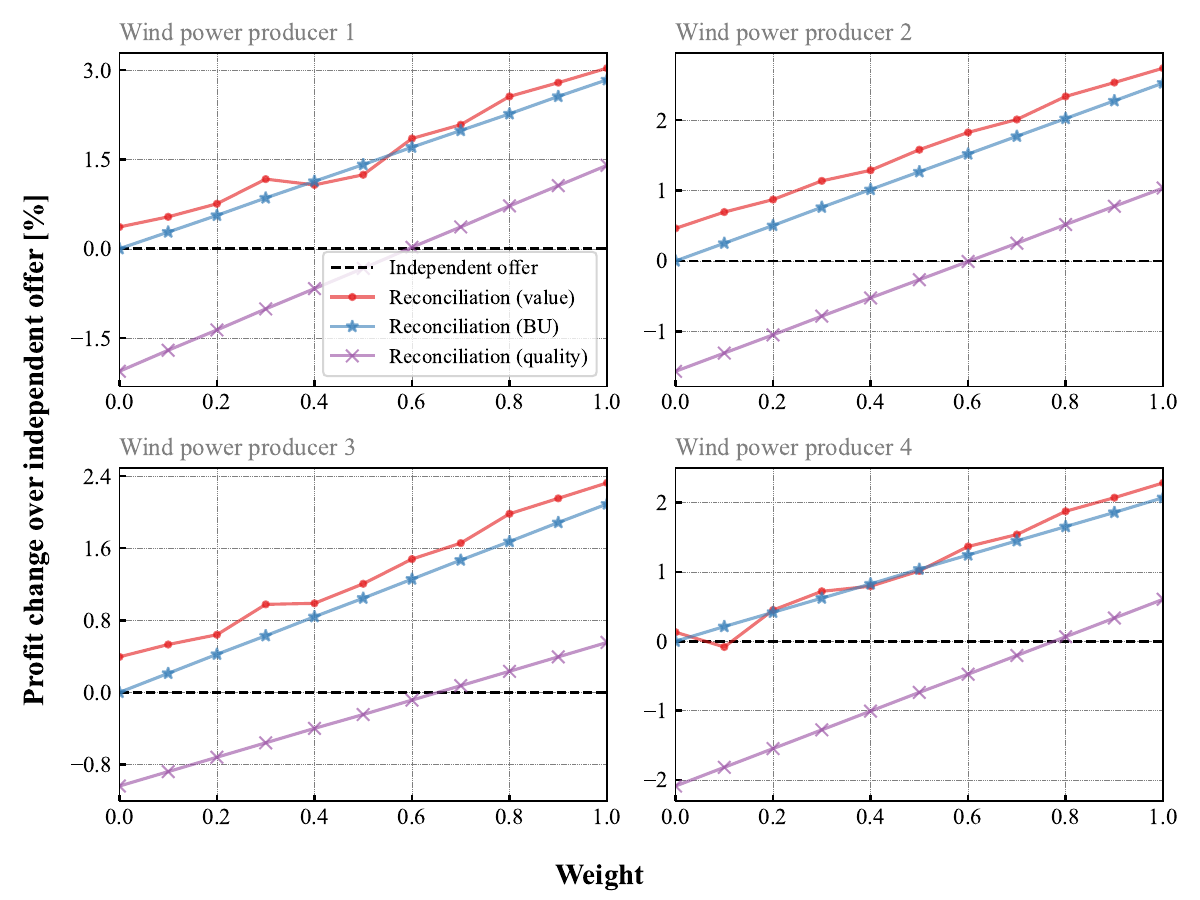}
    \caption{Average profit change over independent offers for each wind power producer across different models in Case 2}
    \label{fig:case 2 av_profit}
\end{figure}

\subsection{Case 2}

We then analyze the average profit of each WPP when base forecasts are communicated as quantiles. The analysis considers varying the weight $w$ in the allocation rule from 0 to 1, with the results shown in Figure~\ref{fig:case 2 av_profit}. Similar to Case 1, the average profits for each WPP increase as $w$ rises, although there are a few exceptions under the value-oriented forecast reconciliation approach. Provided that the unit balancing cost condition (\ref{unit_cost cond}) is satisfied, the profits from BU reconciliation consistently exceed those from independent offering. For quality-oriented reconciliation, the results indicate that the resulting profits are lower than those achieved through strategic independent offering in most cases. This highlights a key limitation of quality-oriented reconciliation when applied in a decision-making context. For the proposed approach, WPP $4$ is an exception when $w=0.1$, primarily due to the challenges in model training at lower values of $w$. In fact, the base forecast profits in this case are substantial, suggesting that the agreement set during bargaining is minimal, which in turn complicates the training process. With more training samples, it is expected to attain estimated parameters within the agreement set. Nevertheless, the profits from the value-oriented forecast reconciliation approach tend to exceed those from BU reconciliation, owing to the information gain discussed in Section~\ref{s4.2.2}.

\begin{table}[htbp]
\centering
\footnotesize
\caption{RMSE of hierarchical forecasts in Cases 1 and 2 (MW)}
\label{tab:rmse_cases}
\begin{tabular}{ll*{10}{c}}
\toprule
 & & \multicolumn{10}{c}{Weight} \\
\cmidrule(lr){3-12}
Case & Method 
& 0.1 & 0.2 & 0.3 & 0.4 & 0.5 & 0.6 & 0.7 & 0.8 & 0.9 & 1.0 \\
\midrule
\multirow{2}{*}{1} 
 & Reconciliation (value)   
 & 0.698 & 0.685 & 0.672 & 0.685 & 0.681 & 0.676 & 0.668 & 0.668 & 0.666 & 0.878 \\
 & Reconciliation (quality) 
 & 0.605 & 0.598 & 0.605 & 0.605 & 0.599 & 0.608 & 0.610 & 0.610 & 0.598 & 0.607 \\
\midrule
\multirow{2}{*}{2} 
 & Reconciliation (value)   
 & 0.689 & 0.695 & 0.692 & 0.689 & 0.689 & 0.688 & 0.682 & 0.679 & 0.674 & 0.806 \\
 & Reconciliation (quality) 
 & 0.601 & 0.605 & 0.602 & 0.604 & 0.603 & 0.600 & 0.608 & 0.612 & 0.606 & 0.602 \\
\bottomrule
\end{tabular}
\end{table}

\begin{figure}[h]
    \centering
    \includegraphics[width=0.9\linewidth]{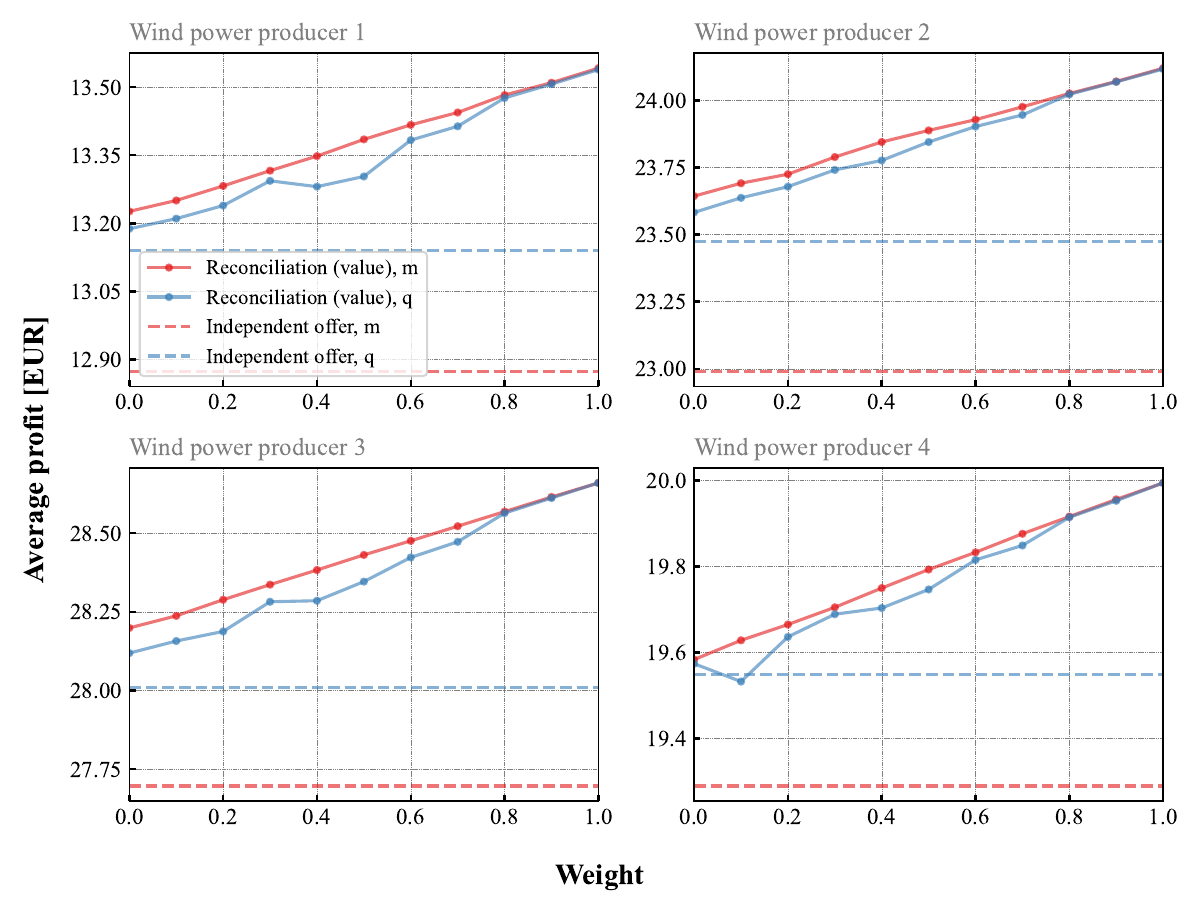}
    \caption{Average profits for each wind power producer in cases where base forecasts are either mean or quantile forecasts}
    \label{fig:m_vs_q av_profit}
\end{figure}

We compare the absolute average profits resulting from the proposed value-oriented forecast reconciliation approach in both scenarios: when the base forecasts are either mean forecasts or quantile forecasts. The results are presented in Figure~\ref{fig:m_vs_q av_profit}, where the `m' and `q' in the legend indicate that the base forecasts are mean or quantile. The analysis shows that, when the base forecasts are mean forecasts, the profits generated by the proposed method exceed those of the alternative approach, even though profits under independent offering remain lower. Notably, when $w$ is either $0.9$ or $1.0$, the profits obtained with the proposed method are almost identical in both scenarios. This finding reinforces the earlier discussion by suggesting that the agreement set becomes more restrictive when base forecasts are quantile forecasts, thereby complicating model training for smaller values of $w$. In contrast, this implies that wind power producers need only to provide their mean forecasts to the PM, which reduces their need to analyze the electricity markets. Furthermore, we report the RMSE of hierarchical forecasts for both cases in Tables~\ref{tab:rmse_cases}. As observed, the forecast errors under the value-oriented reconciliation approach are consistently higher than those obtained from the quality-oriented reconciliation. When considered alongside the profit analysis, these results support our central claim: minimizing overall forecast errors does not necessarily lead to cost reductions for individual agents. Specifically, when the weight $w$ is set to 1.0, the RMSE under the value-oriented reconciliation approach is relatively high, as the loss function exclusively prioritizes the aggregate level.

\section{Real-world Case Study}
\label{s6}

\begin{table}[htbp]
\centering
\footnotesize
\caption{Average profits for each wind power producer under different models [\euro/hour]}
\label{tab:2}
\begin{tabular}{lcccc}
\toprule
\textbf{Model} & \textbf{WPP 1} & \textbf{WPP 2} & \textbf{WPP 3} & \textbf{WPP 4} \\
\midrule
Independent offering 
  & 15.69 & 28.05 & 33.54 & 23.35 \\
BU reconciliation 
  & 15.81 & 28.25 & 33.71 & 23.43 \\
Quality-oriented reconciliation 
  & $15.83 \pm 0.006$ & $28.27 \pm 0.012$ & $33.71 \pm 0.016$ & $23.45 \pm 0.011$ \\
Value-oriented reconciliation 
  & $15.92 \pm 0.001$ & $28.45 \pm 0.004$ & $33.95 \pm 0.003$ & $23.62 \pm 0.002$ \\
\bottomrule
\end{tabular}
\end{table}

Similar to Section \ref{s5}, we also have an hourly forward market followed by a balancing market with a dual-price settlement mechanism. However, unlike the simulation studies, we obtain two years of hourly price data from the Danish TSO data portal\footnote{https://www.energidataservice.dk/}. The hourly penalties are calculated using the day-ahead spot and regulation prices, as specified in Section (\ref{s3}). We consider the four models described in Section \ref{s5}. As noted in the previous section, the mean forecasts suffice as the base forecasts; therefore, we define the base forecasts as the mean forecasts. Specifically, the quality-oriented reconciliation model incorporates lagged wind power generation values as contextual features. While for the value-oriented reconciliation model, we include lagged penalties $\psi^+_{t-h},\cdots,\psi^+_{t}$, $\psi^-_{t-h},\cdots,\psi^-_{t}$ besides wind power generation values as contextual features. In particular, the weight $w$ in the allocation rule is set to $0.9$. For both quality-oriented and value-oriented reconciliation, we repeat the experiments 10 times and report both the mean and standard deviation of the results.

We present the average profits for each wind power producer in Table \ref{tab:2}. The aggregate offering models generate higher average profits compared to independent offering, aligning with the findings from simulation studies. Specifically, BU reconciliation and quality-oriented reconciliation exhibit similar performance, although the latter provides more accurate forecasts. As a result, the profit increase is primarily attributed to participation in aggregate offering. This supports our argument that quality-oriented reconciliation may not be the most effective approach for evaluating forecast accuracy. As expected, value-oriented reconciliation yields the highest average profits for each wind power producer, as it facilitates the development of a profitable trading strategy.

\section{Concluding Remarks}
\label{s7}

Although value is considered a critical factor in evaluating forecasts, alongside quality \citep{murphy1993good}, it has been largely overlooked in the forecast reconciliation literature. Generally, forecast reconciliation can be viewed as an information pool, where a group of agents pools their information to make collective decisions. We emphasize the importance of integrating forecast value into the reconciliation process, demonstrating that quality-oriented approaches may fail to produce decisions that are fair and acceptable to all agents. This issue becomes particularly pronounced when agents with heterogeneous loss functions are involved, as quality-oriented reconciliation may lead to individual revenue losses, which in turn can cause potential conflicts. Consequently, we propose a value-oriented forecast reconciliation approach based on the Nash bargaining framework. Specifically, we frame the problem as a learning-based reconciliation problem, where parameters are estimated using a primal-dual algorithm.

We introduce the concept of consensus on information, distinguishing it from the well-established notion of consensus on allocations, which serves as a prerequisite for multi-agent cooperation. We demonstrate the effectiveness of the proposed value-oriented reconciliation approach through an aggregated wind energy trading problem, where profits are distributed among wind power producers according to a weighted proportional allocation rule. Furthermore, we prove that by employing this allocation rule, the agreement set of parameters within the Nash bargaining game is guaranteed to be non-empty. Given the estimated reconciliation and allocation mechanism, it is challenging for individual WPPs to explicitly derive the functional form associated with their optimal offers. Instead, our framework aggregates the available information and reconciles the forecasts to determine an optimal offer for each WPP in a data-driven manner. Since the reconciled forecasts are directly used as decisions in the market, this approach may also be interpreted as a form of \textit{decision reconciliation}. Specifically, we design simulations and case studies that consider an hourly forward market and a balancing market with a dual-price mechanism, utilizing market data from Denmark and wind power data from the Wind Toolkit. The results show that, under specific conditions, the proposed method ensures that the profits achieved through the aggregate offering approach consistently outperform those obtained through independent offering. This gain results from both information pooling and cooperative trading. Notably, the findings suggest that, in terms of subsequent decision-making value, quality-focused forecast reconciliation may not be the optimal choice.

However, we note that the allocation rule employed in this study does not lie within the core of the cooperative trading game. Furthermore, the proposed rule does not ensure that existing agents in the coalition remain unaffected by the inclusion of new members—an essential property referred to as the population monotonic allocation scheme (PMAS) \citep{sprumont1990population}.Future work should aim to develop allocation mechanisms that both lie within the core and support coalition expansion without disadvantaging any participant. Indeed, the proposed framework is not limited to energy trading; it is also applicable to various domains in supply chain management where cooperation among agents is essential, such as in retail operations. Although this study focuses primarily on point forecast reconciliation, the proposed framework is readily adaptable to broader group decision-making scenarios involving information pooling. Related algorithms may also be designed for reconciling probabilistic forecasts or aggregating multiple forecast sources. Another promising avenue for future research is the incorporation of fairness principles into value-oriented forecasting in more general and diverse application domains.

\section*{Appendix}
\subsection*{Proof of Proposition \ref{theorem4.1}}
\begin{proof}
Using the definition of balancing cost,
\begin{align*}
    c_{t+k}(\Tilde{y}_{t+k,\mathrm{sum}|t},y_{t+k,\mathrm{sum}})&=\psi^+_{t+k}\left[y_{t+k,\mathrm{sum}}-\Tilde{y}_{t+k,\mathrm{sum}|t}\right]^++\psi^-_{t+k}\left[\Tilde{y}_{t+k,\mathrm{sum}|t}-y_{t+k,\mathrm{sum}}\right]^+\\
    &\leq\psi^+_{t+k}\sum_{i=1}^m\left[y_{t+k,i}-\Tilde{y}_{t+k,i|t}\right]^++\psi^-_{t+k}\sum_{i=1}^m\left[\Tilde{y}_{t+k,i|t}-y_{t+k,i}\right]^+\\
    &=\sum_{i=1}^mc_{t+k}(\Tilde{y}_{t+k,i|t},y_{t+k,i}).
\end{align*}
where the inequality results from the sub-additivity property of $[x]^+$ for any $x\in \mathbb{R}$.
\end{proof}

\subsection*{Proof of Proposition \ref{P4.1}}
\begin{proof}
Using the definition of $c_{t+k,i}^{(\mathrm{AG})}(\Tilde{\mathbf{y}}_{t+k|t},\mathbf{y}_{t+k})$ and $\gamma_i^\mathrm{pc}$, we have
\begin{align*}
    c_{t+k,i}^{(\mathrm{AG})}(\Tilde{\mathbf{y}}_{t+k|t},\mathbf{y}_{t+k})&=(1-w)c_{t+k}(\Tilde{y}_{t+k,i|t},y_{t+k,i})+w\frac{c_{t+k}(\Tilde{y}_{t+k,i|t},y_{t+k,i})c_{t+k}(\Tilde{y}_{t+k,\mathrm{sum}|t},y_{t+k,\mathrm{sum}})}{\sum_{i=1}^mc_{t+k}(\Tilde{y}_{t+k,i|t},y_{t+k,i})} \\
    &\leq (1-w)c_{t+k}(\Tilde{y}_{t+k,i|t},y_{t+k,i})+wc_{t+k}(\Tilde{y}_{t+k,i|t},y_{t+k,i})=c_{t+k}(\Tilde{y}_{t+k,i|t},y_{t+k,i}),
\end{align*}
where the inequality results from Proposition \ref{theorem4.1}.
\end{proof}
\subsection*{Batch form of the Lagrangian function}
 Using a batch of samples $\{\hat{\mathbf{y}}_{t+k|t},\mathbf{z}_t,\mathbf{y}_{t+k} \}_{t\in \mathcal{T}_b}$ where $\mathcal{T}_b$ represents the set of indices, we write the Lagrangian of batch samples as
\begin{equation}
\begin{split}
L^{(b)}(\boldsymbol{\theta},\boldsymbol{\mu})&=-\sum_i^m \log\left[\frac{1}{|\mathcal{T}_b|}\sum_{t\in \mathcal{T}_b}\left[c_{t+k}(\hat{y}_{t+k,i|t},y_{t+k,i})-c_{t+k,i}^{(\mathrm{AG})}(\Tilde{\mathbf{y}}_{t+k|t},\mathbf{y}_{t+k})\right] \right]\\
&+\sum_i^m\mu_i\left[\frac{1}{|\mathcal{T}_b|}\sum_{t\in \mathcal{T}_b}\left[c_{t+k,i}^{(\mathrm{AG})}(\Tilde{\mathbf{y}}_{t+k|t},\mathbf{y}_{t+k})-c_{t+k}(\hat{y}_{t+k,i|t},y_{t+k,i})\right] \right]^+,
\end{split}
\end{equation}
where $|\mathcal{T}_b|$ represents the cardinality of $\mathcal{T}_b$.

\section*{Acknowledgment}
We are grateful for the detailed and constructive suggestions from three anonymous reviewers and the editor.
Honglin Wen is funded by the National Science Foundation of China (52307119). Pierre Pinson acknowledges the support of UKRI through the Global NSF-UKRI Centre EPICS (Electric Power Innovation for a Carbon-free Society Centre -- EPICS-UK; grant no. EP/Y025946/1).

\bibliographystyle{apalike} 
\bibliography{mylib}

\end{document}